\pdfoutput=1

\documentclass[11pt]{article}

\usepackage[final]{acl}

\usepackage{times}
\usepackage{latexsym}
\usepackage{bm}
\usepackage{pifont}

\usepackage[T1]{fontenc}

\usepackage[utf8]{inputenc}

\usepackage{microtype}

\usepackage{inconsolata}

\usepackage{graphicx}
\usepackage{subcaption} 
\usepackage{afterpage}
\usepackage{amsmath}
\usepackage{amssymb}
\usepackage{cleveref}
\usepackage{booktabs}
\usepackage{multirow}
\usepackage{siunitx}
\usepackage{float}
\usepackage{stfloats}
\usepackage{makecell}
\usepackage{colortbl}
\usepackage{xcolor} 

\usepackage{graphicx}

\title{EAC-MoE: Expert-Selection Aware Compressor for Mixture-of-Experts Large Language Models}

\author{
Yuanteng Chen$^{1,2}$ \thanks{$\ $  Equal contribution.},
Yuantian Shao$^{3,1}$ \footnotemark[1],
{Peisong Wang}$^{1,2,4}$\thanks{$\ $  Corresponding author.},
\textbf{Jian Cheng}$^{1,2,4,5}$\\
$^1$ $\text{C}^2$DL, Institute of Automation, Chinese Academy of Sciences \\
$^2$School of Artificial Intelligence, University of Chinese Academy of Sciences\\ 
$^3$Nanjing University of Science and Technology \\
$^4$AIRIA  $^5$Maicro.ai  \\
\texttt{chenyuanteng2024@ia.ac.cn, yuantianshao@njust.edu.cn}, \\
\texttt{\{peisong.wang,jcheng\}@nlpr.ia.ac.cn}
}

\begin{document}
\maketitle
\begin{abstract}

Mixture-of-Experts (MoE) has demonstrated promising potential in scaling LLMs. However, it is hindered by two critical challenges: (1) substantial GPU memory consumption to load all experts; (2) low activated parameters cannot be equivalently translated into inference acceleration effects.
In this work, we propose \textbf{EAC-MoE}, an \textbf{E}xpert-Selection \textbf{A}ware \textbf{C}ompressor for \textbf{MoE}-LLMs, which deeply aligns with the characteristics of MoE from the perspectives of quantization and pruning, and introduces two modules to address these two challenges respectively: (1) The expert selection bias caused by low-bit quantization is a major factor contributing to the performance degradation in MoE-LLMs. Based on this, we propose \textbf{Quantization with Expert-Selection Calibration (QESC)}, which mitigates the expert selection bias by calibrating the routers within the MoE; (2) There are always certain experts that are not crucial for the corresponding tasks,  yet causing inference latency. Therefore, we propose \textbf{Pruning based on Expert-Selection Frequency (PESF)}, which significantly improves inference speed by pruning less frequently used experts for current task. Extensive experiments demonstrate that our approach significantly reduces memory usage and improves inference speed with minimal performance degradation.
\end{abstract}


\section{Introduction}

Large Language Models (LLMs) have demonstrated remarkable capabilities in various natural language processing tasks.  \citep{DBLP:conf/iclr/ZhouWLSLQLJSZ024}. 
A recent significant breakthrough in this field is the introduction of the Mixture-of-Experts (MoE) architectures \citep{shazeer2017outrageously, anonymous2024olmoe}. By utilizing a sparse architecture that activates a subset of experts via a dynamic routing mechanism tailored to each input, MoE enables efficient computation and scalable network capacity, matching or exceeding the performance of dense LLMs with several times more activated parameters. 

\begin{figure}[t]
    \centering
    \includegraphics[width=\linewidth]{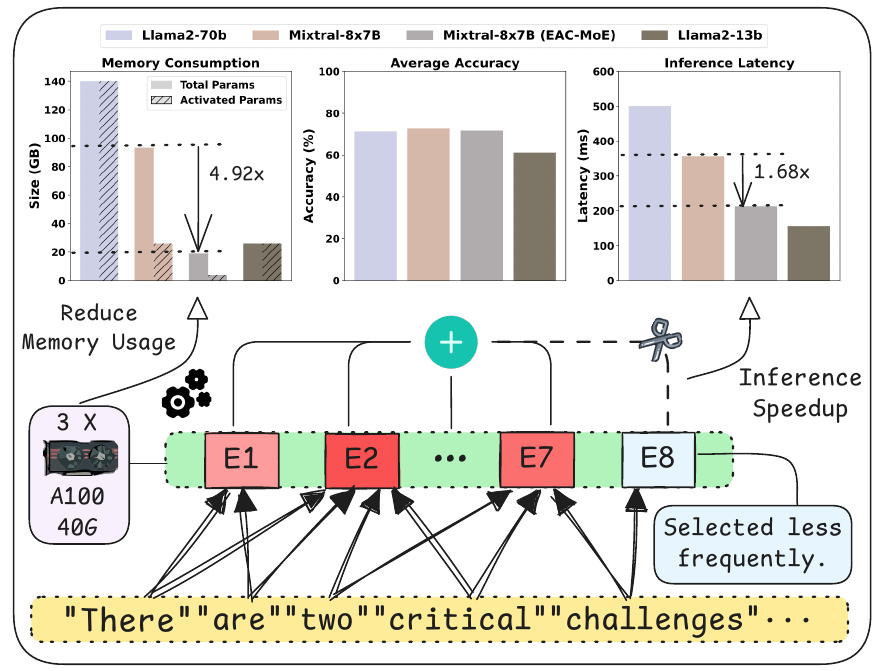}
    \caption{Comprehensive performance of EAC-MoE in reducing memory usage, maintaining model accuracy, and improving inference speed for Mixtral-8x7B. The average accuracy is measured across zero-shot tasks.}
    \label{fig:main_figure}
    \vskip -0.2in
\end{figure}

Although MoE reduces the number of activated parameters through an expert selection mechanism, it does not decrease the total number of model parameters. During inference, all expert weights must be stored in GPU memory, resulting in substantial memory pressure. 
As shown in ~\Cref{fig:main_figure} top, while Mixtral-8x7B \citep{jiang2024mixtral} has a similar activated parameter count to LLaMA2-13B \citep{touvron2023llama}, its total parameter count is about four times larger, occupying 94GB of GPU memory.


On the other hand, the reduction in activated parameters does not directly result in an equivalent speedup during inference. 
Although only a subset of experts is selected for each token, in typical long-sequence or batch inference scenarios, different tokens choose different experts. As illustrated in ~\Cref{fig:main_figure} bottom, MoE still requires computing the output of each expert (E1-E8) separately and performing a weighted summation to obtain the final result, experts like E8 are selected less frequently but still cause non-negligible latency. 

These challenges hinder the practical deployment of MoE models in resource-constrained, low-latency applications. For dense LLMs, quantization and pruning are commonly employed to address these issues. However, directly applying commonly used quantization methods (such as RTN and GPTQ \citep{frantar-gptq}) and pruning methods designed for dense LLMs to MoE models, without considering the characteristics of MoE models, results in significant performance degradation or brings negligible inference speedup. In this work, we design a method that combines quantization and sparse inference, leveraging the expert selection characteristics of MoE models.

In MoE models, the experts are trained to specialize for different types of tasks, and the router can select the most suitable experts for each token, which is the key for its success \citep{6796382}.
However, low-bit quantization of MoE model can bias expert selection probability and cause the router to choose the wrong experts, which we refer to as \textbf{the expert-shift problem}. 
To address this issue in MoE quantization, we propose Quantizaion with Expert-Selection Calibration (QESC): a layer-by-layer router calibration method to mitigate the bias caused by quantization, thereby reducing the shift in expert selection. This approach effectively preserves the performance of the quantized model.

In contrast, the focus of dynamic pruning lies in skipping experts that are relatively unimportant for the current input during inference.
Specifically, certain experts are less frequently selected during inference and have minimal impact on overall performance. Notably, these relatively unimportant experts vary across different types of tasks.
Based on this observation, we propose Pruning based on Expert-Selection Frequency (PESF): a dynamic expert pruning method that prunes less frequently selected experts during inference, 
significantly improving the inference speed of MoE models with minimal performance loss.

Combining QESC and PESF, we propose \textbf{EAC-MoE}, 
exploring the compression of MoE models from both aspects of pre-inference and during-inference. Experiments on four MoE models demonstrate that our method significantly reduces memory usage and improves inference speed. When compressing Mixtral-8x7B, as shown in ~\Cref{fig:main_figure} top, we reduce the memory requirements by 4.92×, enabling deployment on a RTX 3090 GPU. 
Meanwhile, our method achieve 1.68× inference speedups with an average accuracy loss of less than 1\% under simultaneous quantization and pruning, making it practical for real-world applications.

\section{Related Work}
\noindent\textbf{Quantization for LLMs and MoE-LLMs.}
Post-Training Quantization (PTQ) is an efficient technique that reduces computational and storage requirements by converting pre-trained models from high-precision to lower-precision formats without requiring extensive retraining. 
Methods like GPTQ \citep{frantar-gptq} and BiLLM \citep{pmlr-v235-huang24q} focus on addressing weight-only quantization, while approaches such as SmoothQuant \citep{pmlr-v202-xiao23c} and OmniQuant \citep{shao2023omniquant} aim to tackle the challenges of both weight and activation quantization.
In this work, we focus primarily on weight-only quantization because the MoE deployment challenges stem mainly from the memory pressure caused by weight parameters.
For MoE-LLMs, previous studies have largely focused on mixed-precision quantization strategies based on expert selection frequency \citep{li2024examining, huang2024mc}.
Although these methods have shown certain effectiveness, they may face challenges in generalization and risk overfitting.

\noindent\textbf{Pruning of LLMs and MoE-LLMs.} 
Post-training pruning is another key technique to compress LLMs by reducing model size by selectively removing less important parameters while preserving performance \citep{han2015deep, h.2018to, ashkboos2024slicegpt}.
For MoE-LLMs, prior efforts have focused mainly on two directions: pruning experts with lower selection frequency before inference \citep{lu-etal-2024-experts, kim2021scalable}, and pruning less significant weights for each token among the selected experts \citep{lu-etal-2024-experts, huang2024mc}. However, while these approaches have made notable progress, there remain opportunities for further improvement. The first direction, for example, can lead to performance degradation in certain types of tasks. 
The second direction, on the other hand, achieves a relatively low pruning rate, resulting in limited inference speedup.

\begin{figure*}[htb]
    \centering
    \includegraphics[width=\textwidth]{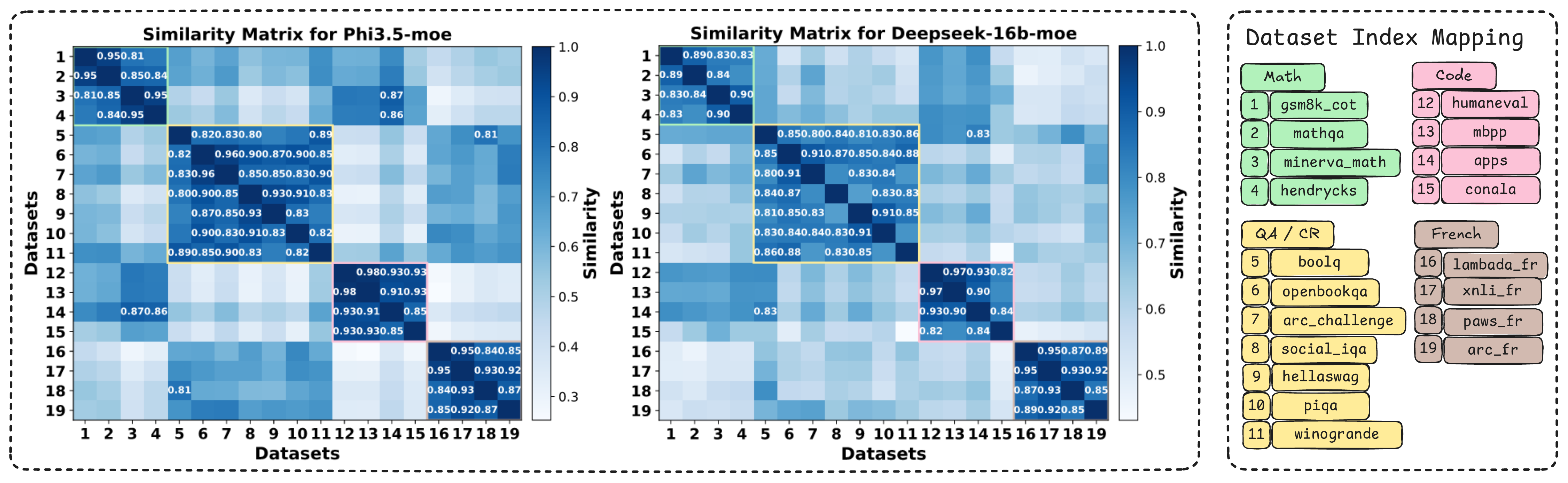}
    \caption{The figure illustrates the pairwise cosine similarity of expert selection frequencies for Phi3.5-moe (left) and Deepseek-moe-16b-base (right) across 19 datasets, which are categorized into four groups distinguished by different colors. Points with cosine similarity greater than 0.8 are highlighted to emphasize high similarity regions.}
    \label{fig:analysis}
    \vskip -0.15in
\end{figure*}

\section{Preliminaries and Motivation}

\subsection{LLM Quantizaiton}
In this work, quantization techniques are employed to compress the weights. Specifically, floating-point weights distributed in \([W_{\text{min}}, W_{\text{max}}]\) are mapped to the integer range \([0, 1, \cdots, 2^B-1]\), where \(B\) represents the target bit-width. The quantization reconstruction problem for the weights \(\bm{W} \in \mathbb{R}^{n_\text{in} \times n_\text{out}}\) can be formulated as:
\begin{equation}
\arg\min_{\bm{W}_q} \|\bm{W}\bm{X} - \bm{W}_q\bm{X}\|_2^2,
\end{equation}
where \(\bm{W}_q\) denotes the quantized weight, 
and \(\bm{X}\) is the input to the layer derived from a small subset of calibration data. 
GPTQ \citep{frantar-gptq} is currently a mainstream weight quantization method, which can efficiently reduce group-wise quantization error by employing Hessian-based estimation (\(\bm{H} = 2\bm{X}\bm{X}^\top\)) and error compensation techniques. It is utilized in subsequent sections of this paper. 

\subsection{Mixture-of-Experts}
Decoder-only MoE models \citep{MLSYS2023_5a54f793} are based on a transformer architecture \citep{10.5555/3295222.3295349}, but the FeedForward Network (FFN) sub-layers of traditional dense models are replaced with MoE layers, each containing $N$ experts. For each input token $\bm{x}$, the router computes routing logits \( \bm{r} = \{r_0, \cdots, r_{N-1}\} \) and expert selection scores \( \bm{s} = \text{Softmax}(\bm{r}) \). The top-$K$ experts are selected based on \( \bm{s} \), and their outputs \( E_{e_j}(\bm{x}) \) are combined as a weighted sum, with normalized weights:
\begin{equation}
\bm{z} = \sum_{j=0}^{K-1} \frac{\bm{s}_{e_j}}{\sum_{i=0}^{K-1} \bm{s}_{e_i}} \cdot E_{e_j}(\bm{x}).
\end{equation}
Here, \( E_{e_j}(\bm{x}) \) represents the output of the \( j \)-th selected expert for the input token \( \bm{x} \). 
Based on this structure and mechanism, models such as Mixtral-8x7B \citep{jiang2024mixtral}, GPT-4 \citep{openai2024gpt4technicalreport} and DeepSeek-V3 \citep{deepseekai2024deepseekv3technicalreport} have achieved superior generative abilities. 

\subsection{Expert-Selection (ES) Analysis}
\label{sec:generalization}
Previous quantization studies for MoE-LLMs have primarily focused on the observation that, during inference, MoE models exhibit significant differences in the selection frequency of different experts \citep{li2024examining}. Consequently, expert selection frequency has been widely adopted as a metric to evaluate the importance of different experts within an MoE layer. However, prior works have overlooked an important pattern: MoE models often demonstrate entirely different expert preferences across different types of tasks. 

To investigate this pattern, we examine three common categories of NLP tasks: Math, Code-Generation, and Question-Answering or Commonsense-Reasoning (QA/CR). Additionally, we analyze tasks in specific languages (French in our case) as a separate category. 
For each dataset, we record the expert selection frequency during inference. Furthermore, we calculate the similarity of expert selection frequencies between every pair of datasets to better understand the diversity in expert preferences across tasks. For a certain MoE layer \( m \) in a MoE model, the normalized expert selection frequency for dataset \( d \) is defined as:
\begin{equation}
P(m, d) = \frac{C(m, d)}{\sum_{i=0}^{N-1} C(m, d, i)}
\end{equation}
where \( C(m, d) = [C(m,d,0), \cdots, C(m, d, N-1)] \), 
with \( C(m, d, i) \) representing the count of the $i$-th expert in layer \( m \) is selected for all input tokens in the dataset \( d \). Then the normalized expert selection frequencies \( P(m, d) \) of all MoE layers are flattened into a single vector \( P(d) \).
Based on this, the similarity of expert preferences between two datasets \( d_i \) and \( d_j \) is computed as:
\begin{equation}
\text{Sim}(d_i, d_j) = \frac{P(d_i) \cdot P(d_j)}{\|P(d_i)\| \|P(d_j)\|}
\end{equation}
As shown in \Cref{fig:analysis}, we calculate the expert preference similarities of Phi3.5-moe \citep{abdin2024phi} and DeepSeek-16b-moe-base models across 19 different datasets. The results indicate that both models reach similar conclusions: expert selection frequencies within datasets of the same task category exhibit high similarity, whereas expert selection frequencies across datasets of different task categories show relatively low similarity.

This observation suggests that MoE models rely primarily on different experts to handle different types of tasks and the importance of the same expert may vary drastically across different tasks, providing us with the following two insights:
\begin{enumerate}
\item For \textbf{static} quantization, we should focus on the expert selection process itself---ensuring that the model can still select the experts important for each task, as we cannot permanently determine the importance of any expert before inference using a calibration set.
\item For \textbf{dynamic} pruning, we should dynamically evaluate the importance of experts based on the type of the current task and prune experts that are not important for the current task.
\end{enumerate}

\section{Quantization with ES Calibration}
\label{sec:method-quantization}
The core idea of our method is to mitigate the performance degradation of quantized MoE models by addressing expert-shift, a critical issue where quantization errors in the multi-head self-attention (MHSA) and MoE blocks distort expert selection probabilities, causing routers to deviate from original expert assignment patterns.
\subsection{Importance of ES Calibration}
We first verify the importance of calibrating expert selection by observing performance degradation caused by expert-shift and performance improvement achieved by preserving the expert selection.
We separately record the expert selection and its corresponding scores ($\bm{s}$) for all inputs on the WikiText2 \citep{merity2016pointer} validation set for both full-precision model and the 3-bit quantized model. Then, we enforce the quantized model to use the expert selection scores of the original precision model for each input (quantized but without expert-shift) and, conversely, enforce the original precision model to use the expert selection scores of the quantized model (not quantized but with expert-shift). Finally, we calculate the perplexity
(PPL) of the inputs under these four conditions respectively.


\begin{table}[t]
\centering
\small
\caption{The impact of weight quantization itself and its induced expert-shift on perplexity (PPL↓) for Mixtral-8x7B and Deepseek-moe-16b-base models.}
\label{tab:quantization_ppl}
\begin{tabular}{lcccc}
\toprule
\textbf{Model} & \textbf{Quantized} & \textbf{Expert-Shift} & \textbf{PPL} \\ 
\midrule
\multirow{4}{*}{Mixtral-8x7B} 
    & \ding{56}      & \ding{56}        & 3.84 \\ 
    & \ding{56}      & \ding{52}        & 4.17 \\
    & \ding{52}      & \ding{56}        & 4.21 \\
    & \ding{52}      & \ding{52}        & 4.65 \\
\midrule
\multirow{4}{*}{\makecell{Deepseek-moe\\-16b-base}} 
    & \ding{56}      & \ding{56}        & 6.51 \\ 
    & \ding{56}      & \ding{52}        & 6.76 \\
    & \ding{52}      & \ding{56}        & 6.81 \\
    & \ding{52}      & \ding{52}        & 7.17 \\
\bottomrule
\end{tabular}
\end{table}

\begin{figure}[t] 
\vskip -0.1in
    \centering
    \includegraphics[width=\linewidth]{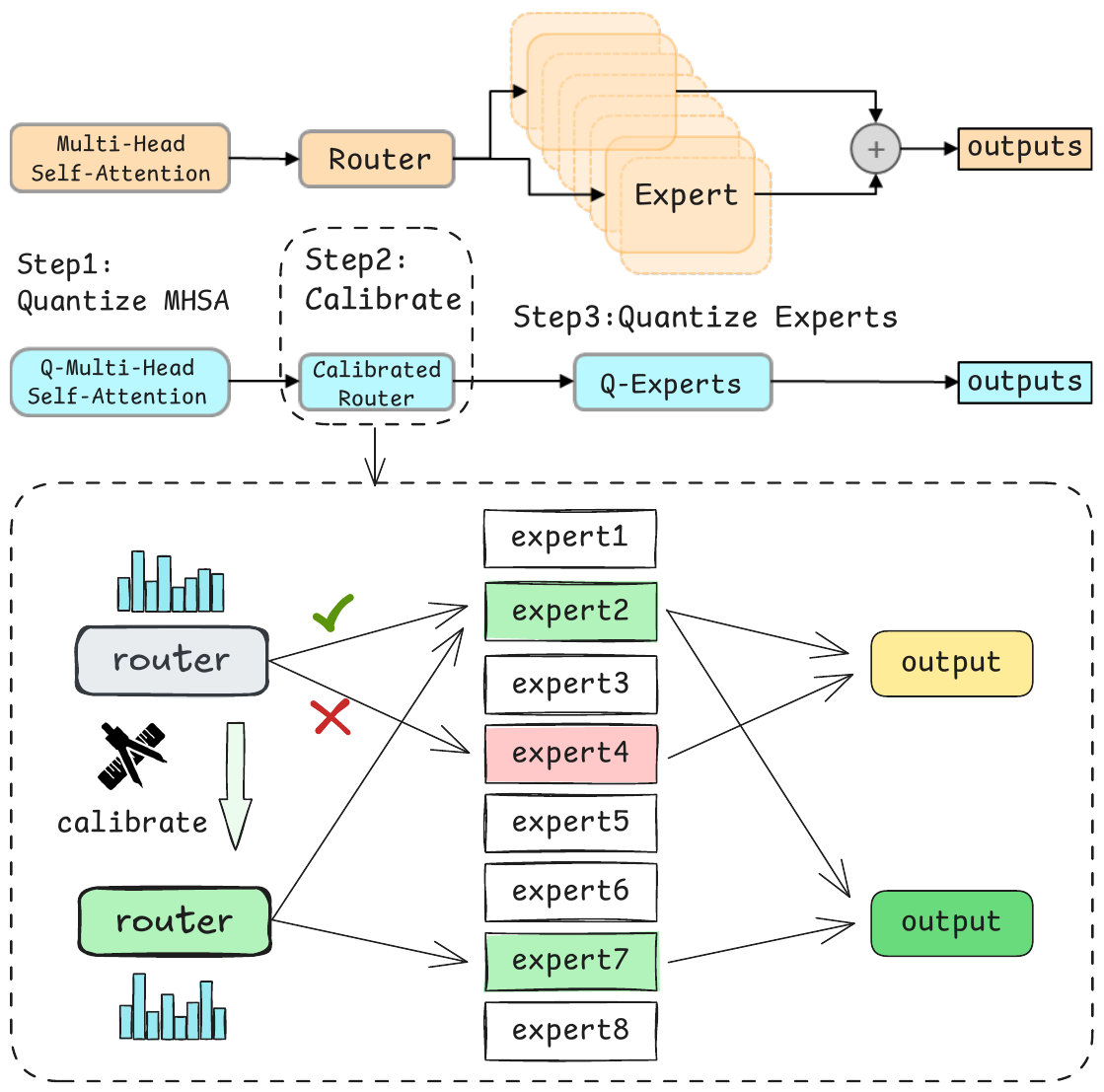}
    \caption{Framework of our proposed Quantization with Experts-Selection Calibration.}
    \label{fig:quant_func}
\vskip -0.1in
\end{figure}

As shown in ~\Cref{tab:quantization_ppl}, expert-shift causes significant performance degradation for the original model. Conversely, preserving the expert selection of the original model significantly improves the performance of quantized models, highlighting the importance of calibrating expert selection.

\subsection{Layer-by-layer Calibration Framework}
Then we focus on how to mitigate expert-shift problem. 
At a hight level, our method performs quantization and calibration layer-by-layer.
Concretely, as illustrated in ~\Cref{fig:quant_func}, using the WikiText2 calibration dataset, we sequentially quantize the MHSA components, calibrate the routers of the MoE layers, and quantize all experts layer by layer.
This process allows the router in each layer to be calibrated in a way that mitigates the expert-shift caused by the quantization of the adjacent layer's MHSA and MoE layer, thereby preventing the cumulative accumulation of expert selection shift across layers.

\begin{figure}[t]
    \centering
    \includegraphics[width=\linewidth]{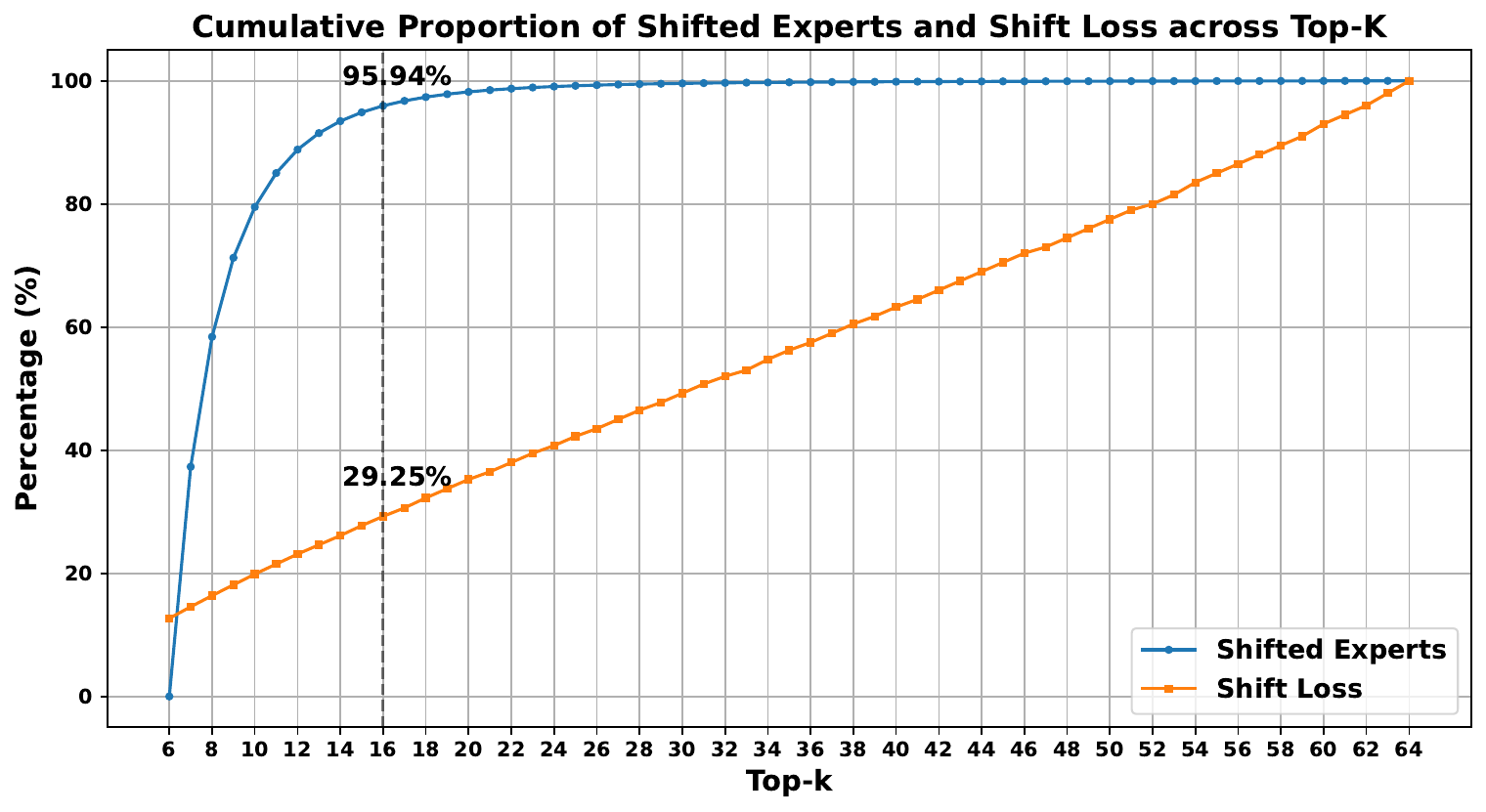}
    \caption{Cumulative proportion of shifted experts in the top-$K$ experts of the probability distribution (blue; i.e., number of shifted experts in top-$K$ / total number of shifted experts), and cumulative proportion of the shift loss of the top-$K$ experts in the probability distribution relative to the total loss of all experts (orange)}
    \label{fig:topk_plot}
\vskip -0.1in
\end{figure}

\subsection{TopK-MSE Loss}
To calibrate the router, a natural idea is to align the router's outputs before and after input quantization, such as by using the mean squared error (MSE) loss for optimization. However, this method is not effective for MoE models with a large number of experts, such as Deepseek-moe-16b-base—which selects 6 experts out of 64 \citep{dai2024deepseekmoe}.
Comparing expert selection before and after 2-bit quantization, as shown in ~\Cref{fig:topk_plot}, we observe that among the experts selected in full precision but not selected after quantization (shifted experts), 95.9\% still rank within the top 16 in the probability distribution. However, the loss corresponding to the top 16 experts accounts for only 29.25\% of the total MSE loss. This indicates that if we directly apply MSE loss to all experts, the loss will be dominated by the majority of experts with very small selection probabilities, which are not selected in full precision, thereby introducing noise into the optimization process. 

Based on this insight,
we adopt the TopK-MSE loss, which computes the MSE loss over only the top-$K$ classes with the highest probabilities, allowing the optimization process to focus on aligning the experts that are more likely to be selected. The TopK-MSE loss is calculated as follows:
\begin{equation}
\mathcal{L} = \frac{1}{K} \sum_{i\in \text{top-$K$}(\bm{W}\bm{x})} \left( (\bm{W}\bm{x})_i - (\bm{W}\bm{\hat{x}})_i \right)^2,
\label{eq:topk_mse_loss}
\end{equation}
$\bm{W}$ represent the weight matrix of router and $\bm{\hat{x}}$ denotes the input obtained from the quantized model. 

\section{Pruning based on ES Frequency}
QESC focuses on ensuring the quantized model can still correctly select the experts important for the current task. A natural consideration is that there are also experts that are not important for the current task. In this section, we introduce a dynamic expert pruning method during inference, which significantly improves inference speed while maintaining almost the same level of accuracy.

Prior work \citep{lu-etal-2024-experts} has already noted the sparsity in expert selection for MoE models, where certain experts are selected with high frequency for a specific task, while others are rarely selected (shown in ~\Cref{sec:sparsity_in_experts}). Meanwhile, as concluded in ~\Cref{sec:generalization}, it is crucial to dynamically evaluate the importance of each expert during inference for different tasks. Therefore, unlike prior work that performs static expert pruning based on selection frequency before inference, our approach dynamically identifies experts that are less important for the current task during the inference process. This allows us to achieve significant inference speedup with minimal performance degradation.

\begin{figure}[ht]
    \centering
    \includegraphics[width=\linewidth]{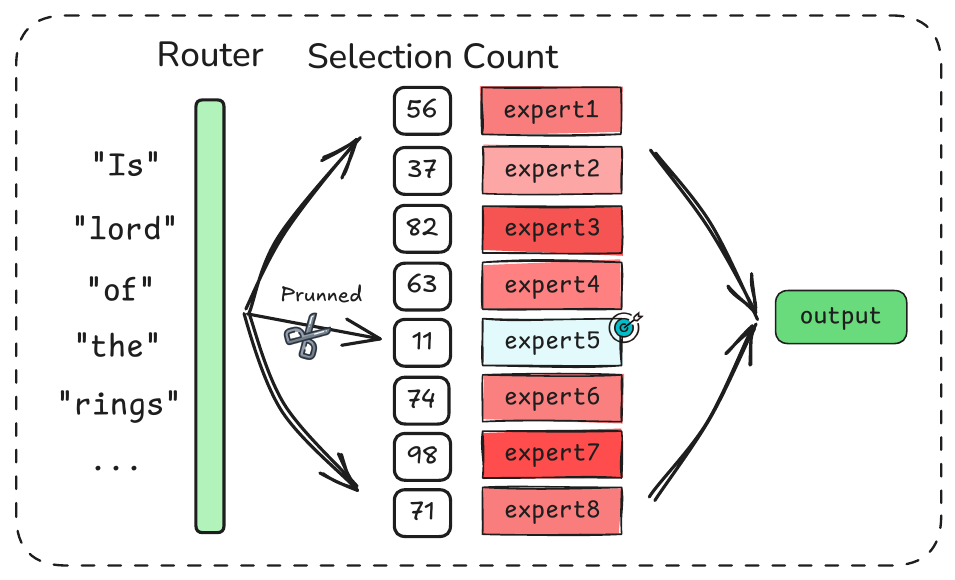}
    \caption{Framework of our proposed Pruning based on Experts-Selection Frequency.}
    \label{fig:pruning}
\end{figure}

In our method, the dynamic pruning criterion is set as follows: assume each layer of the MoE model has \( N \) experts, each token selects \( K \) experts, and the input sequence length is \( l \). The dynamic \textbf{pruning threshold} is defined as \( \alpha \) (\( 0 < \alpha \leq 1 \)). If the number of times an expert is selected, denoted as \( c \), satisfies the condition:
\begin{equation}
c < \left( \frac{l \times K}{N} \right) \times \alpha
\label{eq:pruning_formula}
\end{equation}
then the expert is pruned. In other words, if an expert is selected less frequently than the average selected count multiplied by the threshold \( \alpha \) (like \(expert5\) in ~\Cref{fig:pruning}), it is pruned and excluded from the computation for this sequence.


\begin{table*}[t]
    \vskip -0.1in
    \caption{Comparison of the average perplexity (PPL) scores on WikiText2 validation set and the average accuracy on 8 zero-shot tasks across four different MoE models. We reproduce results of BSP and PMQ on four models using the official codebases provided in their repositories (the reproduction details are provided in ~\Cref{sec:repoduce-quant}) and evaluated all the results under the same settings.  Full results are in the ~\Cref{sec:all-results}.}
    \label{tab:quantization_comparison}
    \centering
    \resizebox{\textwidth}{!}{
    \begin{tabular}{c|c||cc|cc|cc|cc}
        \hline
        \hline
        \multirow{2}{*}{\parbox{2cm}{\centering Bits}} & \multirow{2}{*}{\parbox{3cm}{\centering Method}} & \multicolumn{2}{c|}{Mixtral-8x7B} & \multicolumn{2}{c|}{Phi3.5-moe} & \multicolumn{2}{c|}{Deepseek-moe-16b-base} & \multicolumn{2}{c}{Qwen1.5-MoE-A2.7B} \\ 
        ~ & ~ & PPL $\downarrow$ & $\text{0-shot}^{8} \uparrow$ & PPL $\downarrow$ & $\text{0-shot}^{8} \uparrow$ & PPL $\downarrow$ & $\text{0-shot}^{8} \uparrow$ & PPL $\downarrow$ & $\text{0-shot}^{8} \uparrow$ \\
        \hline
        \multirow{1}{*}{16.00} & Baseline & 3.84 & 72.64 & 3.99 & 69.62 & 6.51 & 61.38 & 7.22 & 64.72 \\ 
        \hline
        \multirow{3}{*}{2.06} 
        & GPTQ & 5.51 & 62.56 & 5.32 & 64.45 & 8.27 & 54.88 & 9.92 & 57.76 \\
        ~ & PMQ & 5.41 & 63.25 & 5.88 & 61.35 & 8.42 & 54.79 & 9.89 & 57.79 \\
        ~ & \cellcolor{gray!15} \textbf{QESC} & \cellcolor{gray!15}\makebox[0pt][c]{\textbf{5.09}} & \cellcolor{gray!15}\makebox[0pt][c]{\textbf{66.31}} & \cellcolor{gray!15}\makebox[0pt][c]{\textbf{5.22}} & \cellcolor{gray!15}\makebox[0pt][c]{\textbf{65.03}} & \cellcolor{gray!15}\makebox[0pt][c]{\textbf{7.99}} & \cellcolor{gray!15}\makebox[0pt][c]{\textbf{57.05}} & \cellcolor{gray!15}\makebox[0pt][c]{\textbf{8.30}} & \cellcolor{gray!15}\makebox[0pt][c]{\textbf{59.52}} \\ 
        \hline
        \multirow{4}{*}{2.54} 
        & GPTQ & 4.74 & 68.65 & 4.74 & 65.81 & 7.36 & 56.83 & 8.41 & 57.91 \\ 
        ~ & BSP & 4.98 & 65.44 & 4.72 & 66.15 & 7.32 & 58.24 & 8.11 & 60.40 \\ 
        ~ & PMQ & 4.78 & 67.5 & 4.73 & 66.03 & 7.17 & 58 & 8.09 & 60.47 \\ 
        ~ & \cellcolor{gray!15} \textbf{QESC} & \cellcolor{gray!15}\makebox[0pt][c]{\textbf{4.54}} & \cellcolor{gray!15}\makebox[0pt][c]{\textbf{69.61}} & \cellcolor{gray!15}\makebox[0pt][c]{\textbf{4.66}} & \cellcolor{gray!15}\makebox[0pt][c]{\textbf{66.53}} & \cellcolor{gray!15}\makebox[0pt][c]{\textbf{7.08}} & \cellcolor{gray!15}\makebox[0pt][c]{\textbf{58.33}} & \cellcolor{gray!15}\makebox[0pt][c]{\textbf{7.74}} & \cellcolor{gray!15}\makebox[0pt][c]{\textbf{61.47}} \\ 
        \hline
        \multirow{3}{*}{3.03} 
        & GPTQ & 4.16 & 68.92 & 4.28 & 68.12 & 6.82 & 59.33 & 7.69 & 62.21 \\
        ~ & BSP & 4.25 & 67.22 & 4.61 & 67.67 & 7.05 & 59.39 & 7.86 & 60.88 \\
        ~ & \cellcolor{gray!15} \textbf{QESC} & \cellcolor{gray!15}\makebox[0pt][c]{\textbf{4.14}} & \cellcolor{gray!15}\makebox[0pt][c]{\textbf{72.21}} & \cellcolor{gray!15}\makebox[0pt][c]{\textbf{4.24}} & \cellcolor{gray!15}\makebox[0pt][c]{\textbf{68.49}} & \cellcolor{gray!15}\makebox[0pt][c]{\textbf{6.71}} & \cellcolor{gray!15}\makebox[0pt][c]{\textbf{61.22}} & \cellcolor{gray!15}\makebox[0pt][c]{\textbf{7.50}} & \cellcolor{gray!15}\makebox[0pt][c]{\textbf{62.89}} \\ 
        \hline
        \hline
    \end{tabular}}
    \vskip -0.1in
\end{table*}

\section{Experiment}
In this section, 
we first evaluate the experimental performance of our proposed methods QESC and PESF, respectively. Then we combine quantization and pruning (QESC+PESF) to assess their performance in maintaining model accuracy, memory usage reduction, and actual inference speedup.

\subsection{Setup}
\noindent\textbf{Models and Dataset.} We validate our method on four MoE models: Mixtral-8x7B, Phi3.5-moe, Deepseek-moe-16b-base and Qwen1.5-MoE-A2.7B \citep{yang2024qwen2}. We report perplexity (PPL) on the WikiText2 testset and accuracies of eight zero-shot tasks tested by EleutherAI LM Harness \citep{eval-harness}, including Winogrande \citep{ai2:winogrande}, PIQA \citep{Bisk2020}, ARC-Easy, ARC-Challenge \citep{allenai:arc}, BoolQ \citep{clark2019boolq}, MathQA \citep{amini-etal-2019-mathqa}, HellaSwag \citep{zellers2019hellaswag}, MMLU \citep{hendrycks2021ethics}. Additionally, we present the results of our method on the challenging tasks GSM8K \citep{cobbe2021gsm8k} and HumanEval \citep{chen2021evaluating}.
\begin{figure}[ht]
    \centering
    \includegraphics[width=\linewidth]{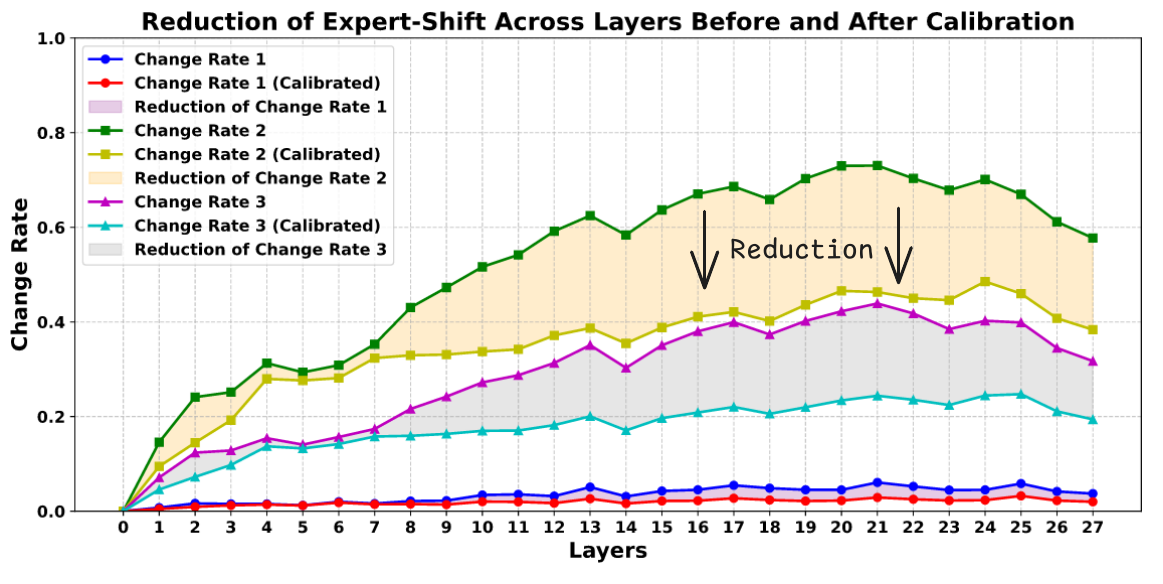}
    \caption{The reduction of expert-shift before and after calibration measured by expert-selection change rate across layers in Deepseek-moe-16b-base under 2.06-bit quantization. Change Rate 1-3 respectively represent three metrics: all expert selections changed, at least one selection changed and half or more selections changed.}
    \label{fig:change_rate}
    \vskip -0.1in
\end{figure}


\noindent\textbf{Implementation Details.}
We follow the settings of prior work \citep{li2024examining, huang2024mc}, keeping the MHSA components at 4-bit precision, while quantizing all experts to 2-bit or 3-bit precision, and maintaining the router at its original precision. Overall, we evaluate our method under three average bit-width settings: 2.06-bit, 2.54-bit, and 3.03-bit (detailed bit-width setting is discussed in ~\Cref{sec:detailed-bit}). The quantization employs group-wise (group size 128) asymmetric quantization and follows the GPTQ procedure. 
We use 128 sequences of length 2048 from the WikiText2 training set as the calibration set for QESC.

\begin{table*}[ht]
    \vskip -0.1in
    \caption{Comparison of the average accuracy on 8 zero-shot tasks and speedup of inference across four different MoE models. The speedup is calculated based on the total inference time of the model with dynamic pruning compared to the original model across 8 tasks. We reproduce results of EES and ODP (details are provided in ~\Cref{sec:repoduce-pruning}) and evaluate all the results under the same settings. Full results can be found in ~\Cref{sec:all-results-pruning}.}
    \label{tab:speedup_comparison}
    \centering
    \resizebox{\textwidth}{!}{
    \begin{tabular}{c||cc|cc|cc|cc}
        \hline
        \hline
        \multirow{2}{*}{\parbox{3cm}{\centering Method}} & \multicolumn{2}{c|}{Mixtral-8x7B} & \multicolumn{2}{c|}{Phi3.5-moe} & \multicolumn{2}{c|}{Deepseek-moe-16b-base} & \multicolumn{2}{c}{Qwen1.5-MoE-A2.7B} \\ 
        ~ & $\text{0-shot} \uparrow$ & Speedup $\uparrow$ & $\text{0-shot} \uparrow$ & Speedup $\uparrow$ & $\text{0-shot} \uparrow$ & Speedup $\uparrow$ & $\text{0-shot} \uparrow$ & Speedup $\uparrow$ \\
        \hline
        Baseline & 72.64 & 1.00 & 69.62 & 1.00 & 61.38 & 1.00 & 64.72 & 1.00 \\ 
        \hline
        EES & 71.40 & 1.06 & 67.96 & 1.05 & 61.15 & 1.08 & 64.42 & 1.06 \\ 
        \hline
        ODP & 71.98 & 1.05 & 68.92 & 1.04 & 61.19 & 1.08 & 64.48 & 1.06 \\ 
        \hline
        \cellcolor{gray!15} PESF (\(\alpha=0.3\)) & \cellcolor{gray!15}\makebox[0pt][c]{\textbf{72.19}} & \cellcolor{gray!15}\makebox[0pt][c]{1.08} & \cellcolor{gray!15}\makebox[0pt][c]{\textbf{69.27}} & \cellcolor{gray!15}\makebox[0pt][c]{1.12} & \cellcolor{gray!15}\makebox[0pt][c]{\textbf{61.28}} & \cellcolor{gray!15}\makebox[0pt][c]{1.11} & \cellcolor{gray!15}\makebox[0pt][c]{\textbf{64.64}} & \cellcolor{gray!15}\makebox[0pt][c]{1.14} \\ 
        \hline
        \cellcolor{gray!30} PESF (\(\alpha=0.7\)) & \cellcolor{gray!30}\makebox[0pt][c]{58.22} & \cellcolor{gray!30}\makebox[0pt][c]{\textbf{1.13}} & \cellcolor{gray!30}\makebox[0pt][c]{67.95} & \cellcolor{gray!30}\makebox[0pt][c]{\textbf{1.30}} & \cellcolor{gray!30}\makebox[0pt][c]{60.41} & \cellcolor{gray!30}\makebox[0pt][c]{\textbf{1.45}} & \cellcolor{gray!30}\makebox[0pt][c]{63.87} & \cellcolor{gray!30}\makebox[0pt][c]{\textbf{1.47}} \\ 
        \hline
        \hline
    \end{tabular}}
    \vskip -0.2in
\end{table*}

\subsection{Experiment on Quantization}
\label{sec:exp_quantization}
\noindent\textbf{Reduction in Expert-Shift.}
First, we intuitively validate the effectiveness of our calibration method by measuring the expert selection change rate before and after calibration on WikiText2 validation set. 
We calculate the expert selection change rates of the quantized model with or without router calibration 
relative to the full-precision model on Deepseek-moe-16b-base, and show the relative reduction in ~\Cref{fig:change_rate}. The results demonstrate that our calibration method significantly reduces the expert selection change rate in quantized MoE models across three metrics.

\noindent\textbf{Overall Performance.} 
We further validate the overall performance of our method.
We compare our quantization method with three other methods: GPTQ, PMQ \citep{li2024examining}, and BSP \citep{li2024examining}. 
GPTQ serves as the baseline for uniform bit-width quantization, while PMQ (1.57–2.54 bit) and BSP (2.54–3.03 bit) are current SOTA methods for mixed-precision quantization of MoE models.
It is worth noting that QESC is inherently orthogonal to other weight quantization approaches for LLMs that focus on minimizing quantization error.

As shown in ~\Cref{tab:quantization_comparison}, when only GPTQ is used to reduce quantization loss, significant performance degradation is still observed. 
Both BSP and PMQ, as mixed-precision quantization methods, demonstrate performance improvements over GPTQ at certain quantization bit-widths for some models. However, in nearly half of the settings, their results are inferior to those of GPTQ, indicating a certain degree of lack of generalization. In contrast, the proposed QESC method significantly outperforms GPTQ, BSP, and PMQ across all results. For instance, at 2.54-bit, QESC limits the performance loss to around 3\% for all four models. Notably, at 3.03-bit, QESC reduces the loss to within 0.5\% for Mixtral-8x7B and Deepseek-moe-16b-base, making it suitable for practical application scenarios.

\noindent\textbf{Challenging Tasks.} Apart from PPL and common-sense tasks, we also evaluate our QESC method on the challenging tasks GSM8K and HumanEval, with the results provided in \Cref{sec:challengeing_tasks}. 




\begin{figure}[htb]
    \centering
    \includegraphics[width=\linewidth]{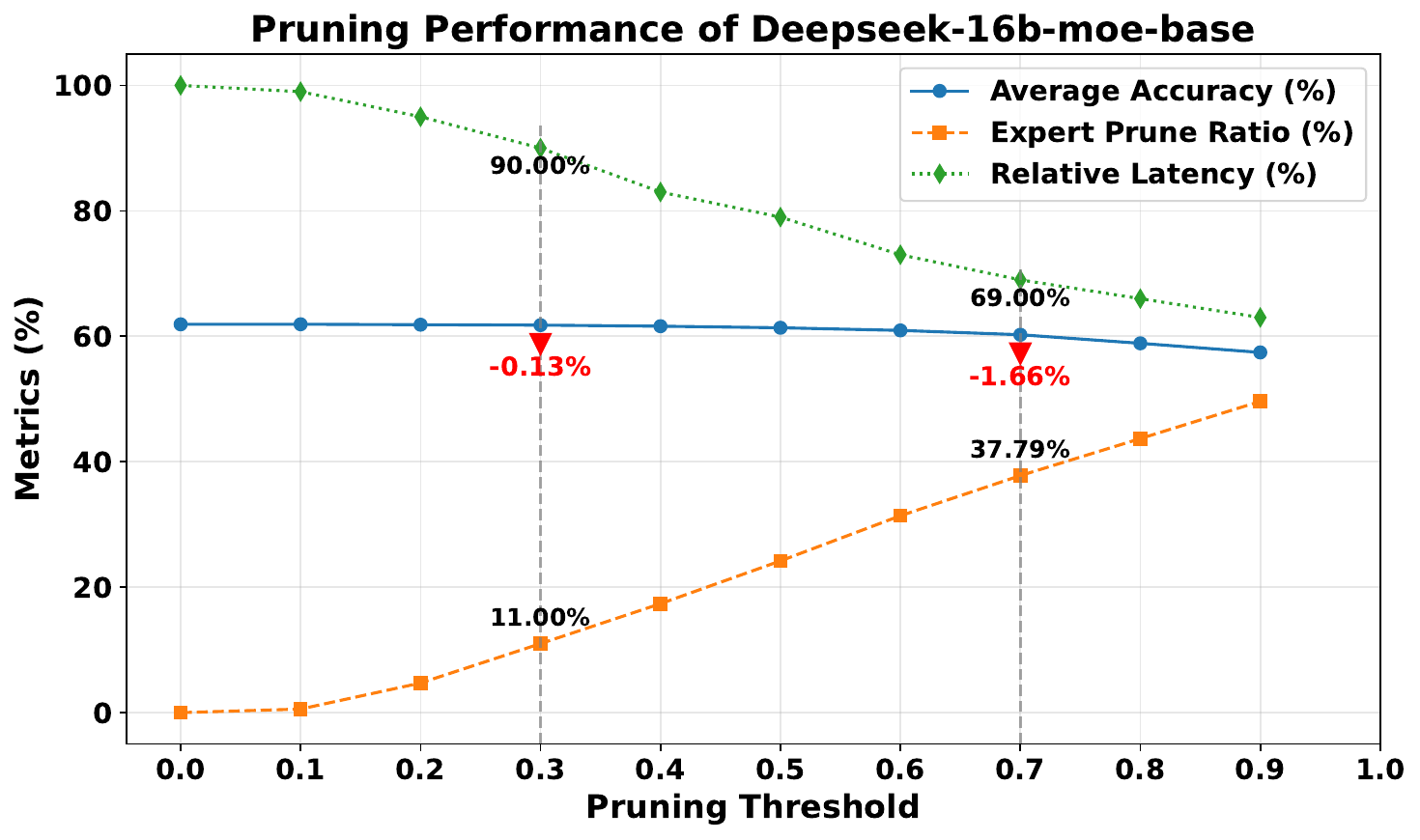}
    \caption{The variations in the model's average accuracy, expert pruning rate, and inference acceleration effect with respect to changes in the pruning threshold \( \alpha \).}
    \label{fig:pruning-result}
    \vskip -0.15in
\end{figure}


\subsection{Experiment on Pruning}
\label{sec:exp-pruning}
\noindent\textbf{Pruning Threshold Analysis.}
To determine a relatively appropriate pruning threshold, we aim to trade off among model accuracy, expert pruning rate, and relative inference latency.  As shown in ~\Cref{fig:pruning-result}, we conduct experiments on Deepseek-16b-moe-base, adjusting the pruning threshold (\(\alpha\)) from 0 to 0.9 with an interval of 0.1. For each threshold, we calculate the average accuracy on 8 zero-shot tasks, the average expert pruning rate across all layers, and the percentage of relative inference latency compared to the original model.
The results show that pruning thresholds of 0.3 and 0.7 represent two sweet spots. The former achieves approximately 10\% speed improvement with almost no loss to the model (average loss within 0.5\%), while the latter is more aggressive, achieving over 1.3$\times$ average inference speedup while still keeping the average loss within around 1.5\%.

\noindent\textbf{Overall Performance.} We compare our method with the classical MoE expert pruning method, known as Efficient Experts Skipping (EES) \citep{lu-etal-2024-experts}, and a recently proposed MoE pruning method, ODP \citep{huang2024mc}. EES performs pruning from the perspective of individual tokens, skipping the selected experts with negligible scores for each input token, while ODP incorporates a key token protection mechanism on top of this. However, both methods can only reduce the input size for a subset of experts, resulting in limited inference speedup. In contrast, our PESF method performs pruning from the perspective of experts, directly skipping experts that are selected less frequently for the current sequence. As shown in ~\Cref{tab:speedup_comparison}, under the more conservative setting (\(\alpha=0.3\)), our method significantly outperforms EES and ODP on all four models in both average accuracy and relative speedup. Moreover, compared to EES and ODP, our pruning method demonstrates greater flexibility. Notably, when we adopt a more aggressive setting with (\(\alpha=0.7\)), except for Mixtral-8x7B (discussed in ~\Cref{sec:pruning-mixtral-8x7B}), our method achieves an inference speedup of 1.30x or greater on the other three models, while still maintaining model accuracy comparable to ODP.

\begin{table}[t]
    \caption{The overall performance of our compression method which combines QESC (3.03 bit) and PESF (\(\alpha=0.3\)). ”Params” denotes the parameter size, including quantizer parameters for the compressed model. 
}
    \label{tab:EAC-MoE-performance}
    \centering
    \resizebox{0.48\textwidth}{!}{ 
    \begin{tabular}{c|c||ccc}
        \hline
        \hline
        \multirow{1}{*}{\parbox{2cm}{\centering Models}} & \multirow{1}{*}{\parbox{2cm}{\centering Method}} & Params(GB) & $\text{0-shot}^{8} \uparrow$ & Speedup $\uparrow$ \\
        \hline
        \multirow{3}{*}{Mixtral-8x7B} 
        & Baseline & 93.41 & 72.64 & 1.00 \\
        & QESC & 18.98 & 72.21 & 1.54 \\
        ~ & \cellcolor{gray!20}\makebox[0pt][c]{QESC+PESF} & \cellcolor{gray!20}\makebox[0pt][c]{18.98} & \cellcolor{gray!20}\makebox[0pt][c]{71.68} & \cellcolor{gray!20}\makebox[0pt][c]{1.68} \\ 
        \hline
        \multirow{3}{*}{Phi3.5-moe} 
        & Baseline & 83.75 & 69.62 & 1.00 \\ 
        & QESC & 17.08 & 68.49 & 1.55 \\
        ~ & \cellcolor{gray!20}\makebox[0pt][c]{QESC+PESF} & \cellcolor{gray!20}\makebox[0pt][c]{17.08} & \cellcolor{gray!20}\makebox[0pt][c]{68.31} & \cellcolor{gray!20}\makebox[0pt][c]{1.75} \\ 
        \hline
        \multirow{3}{*}{\makecell{Deepseek-moe\\-16b-base}} 
        & Baseline & 32.75 & 61.38 & 1.00 \\ 
        & QESC & 7.19 & 61.22 & 1.39 \\
        ~ & \cellcolor{gray!20}\makebox[0pt][c]{QESC+PESF} & \cellcolor{gray!20}\makebox[0pt][c]{7.19} & \cellcolor{gray!20}\makebox[0pt][c]{61.09} & \cellcolor{gray!20}\makebox[0pt][c]{1.55} \\ 
        \hline
        \multirow{3}{*}{\makecell{Qwen1.5-MoE\\-A2.7B}} 
        & Baseline & 28.63 & 64.72 & 1.00 \\ 
        & QESC & 6.69 & 62.89 & 1.36 \\
        ~ & \cellcolor{gray!20}\makebox[0pt][c]{QESC+PESF} & \cellcolor{gray!20}\makebox[0pt][c]{6.69} & \cellcolor{gray!20}\makebox[0pt][c]{62.73} & \cellcolor{gray!20}\makebox[0pt][c]{1.58} \\ 
        \hline
        \hline
    \end{tabular}
    }
    \vskip -0.2in
\end{table}

\subsection{Experiment on Quantization + Pruning}
Finally, we apply our QESC and PESF methods together to comprehensively compress MoE-LLMs.
To achieve a reasonable trade-off between reducing memory usage, inference speed, and maintaining model performance, we apply a relatively mild dynamic pruning strategy (\(\alpha=0.3\)) on top of 3.03-bit static quantization. We report the memory usage, average accuracy on zero-shot tasks, and inference speedup measured by the context latency for a batch of 4 sentences of length 512 in ~\Cref{tab:EAC-MoE-performance}. 

\noindent\textbf{Maintain Accuracy.} With the aid of effective expert selection calibration, our method limits the average accuracy loss across four models to within 1.25\%, effectively maintaining the accuracy of the compressed MoE models.

\noindent\textbf{Memory Saving and Inference Efficiency.} 
By leveraging the BitBLAS tool \citep{ladder-osdi24} to store quantized weights and efficiently handle mixed-precision BLAS operations on GPUs, we limit the memory usage of Mixtral-8x7B and Phi3.5-moe to within 19GB, and that of Deepseek-moe-16b-base and Qwen1.5-MoE-A2.7B to within 7.2GB. This optimization enables deployment on a single RTX 3090 GPU while achieving an average speedup of 1.49$\times$ under 3.03-bit quantization. Furthermore, by integrating efficient dynamic expert pruning, we attain an average actual inference speedup of 1.64$\times$ across all four models.

\noindent\textbf{Comparion with MC-MoE.}
To the best of our knowledge, MC-MoE \citep{huang2024mc} is currently the only method that leverages both static quantization and dynamic pruning for MoE-LLMs, providing specific implementations for 2.06-bit and 2.56-bit quantization and pruning on Mixtral-8x7B. Therefore, we compare our method with MC-MoE at the corresponding quantization bit-widths on the same model and adopt a more conservative pruning strategy in PESF (\(\alpha=0.3\)). As shown in ~\Cref{tab:quant+pruning}, our method outperforms MC-MOE in terms of PPL, average accuracy on zero-shot tasks, and actual inference speedup under both quantization settings.

\begin{table}[t]
    \caption{Comprehensive comparison of average accuracy on 8 zero-shot tasks and inference speedup of four models under quantization and pruning.}
    \label{tab:quant+pruning}
    \centering
    \resizebox{0.48\textwidth}{!}{ 
    \begin{tabular}{c|c||ccc}
        \hline
        \hline
        \multirow{2}{*}{\parbox{1.5cm}{\centering Bits}} & \multirow{2}{*}{\parbox{3cm}{\centering Method}} & \multicolumn{3}{c}{Mixtral-8x7B} \\ 
        ~ & ~ & PPL $\downarrow$ & $\text{0-shot}^{8} \uparrow$ & Speedup $\uparrow$ \\
        \hline
        \multirow{1}{*}{16.00} & Baseline & 3.84 & 72.64 & 1.00 \\ 
        \hline
        \multirow{2}{*}{2.06} 
        & MC-MoE & 5.51 & 62.56 & 1.80 \\
        ~ & \cellcolor{gray!15}\makebox[0pt][c]{\textbf{EAC-MoE (ours)}} & \cellcolor{gray!15}\makebox[0pt][c]{\textbf{5.14}} & \cellcolor{gray!15}\makebox[0pt][c]{\textbf{65.90}} & \cellcolor{gray!15}\makebox[0pt][c]{\textbf{1.82}} \\ 
        \hline
        \multirow{2}{*}{2.56} 
        & MC-MoE & 4.74 & 68.65 & 1.71 \\ 
        ~ & \cellcolor{gray!15}\makebox[0pt][c]{\textbf{EAC-MoE (ours)}} & \cellcolor{gray!15}\makebox[0pt][c]{\textbf{4.58}} & \cellcolor{gray!15}\makebox[0pt][c]{\textbf{68.60}} & \cellcolor{gray!15}\makebox[0pt][c]{\textbf{1.74}} \\ 
        \hline
        \hline
    \end{tabular}
    }
    \vskip -0.2in
\end{table}

\noindent\textbf{More Results of EAC-MoE.} Additionally, we perform more detailed experiments by combining other quantization bit-widths and more aggressive pruning strategies across all four models. Detailed results can be found in ~\Cref{sec:all-results-EAC-MOE}.

\subsection{Ablation Study of Loss Type}
We compare the average accuracy on 0-shot tasks after calibration using TopK-MSE and MSE loss on three MoE models with a larger number of experts (the search for the optimal k-values in shown in ~\Cref{sec:best_k_in_topk}). As shown in ~\Cref{tab:ablation}, the calibrated models optimized with TopK-MSE demonstrate significantly better performance, proving the effectiveness of our optimization method.

\begin{table}[ht]
    \caption{The impact of different loss types on the average accuracy of the calibrated model on 0-shot tasks (under 2.06-bit quantization).
}
    \label{tab:ablation}
    \centering
    \resizebox{0.48\textwidth}{!}{ 
    \begin{tabular}{c|c||cc}
        \hline
        \hline
        \multirow{1}{*}{\parbox{2cm}{\centering Models}} & \multirow{1}{*}{\parbox{2cm}{\centering Loss Type}} & PPL $\downarrow$ & $\text{0-shot}^{8} \uparrow$ \\
        \hline
        \multirow{2}{*}{Phi3.5-moe} 
        & MSE & 5.33 & 64.52 \\ 
        ~ & TopK-MSE & 5.22 & 65.03 \\ 
        \hline
        \multirow{2}{*}{\makecell{Deepseek-moe\\-16b-base}} 
        & MSE & 8.16 & 55.91 \\ 
        ~ & TopK-MSE & 7.99 & 57.05 \\ 
        \hline
        \multirow{2}{*}{\makecell{Qwen1.5-MoE\\-A2.7B}} 
        & MSE & 9.02 & 58.44 \\ 
        ~ & TopK-MSE & \num{8.30} & 59.52 \\ 
        \hline
        \hline
    \end{tabular}
    }
    \vskip -0.2in
\end{table}

\section{Conclusion}
In this work, we aim to address the challenges faced by MoE-LLMs and the limitations of existing compression methods. Focusing on expert selection, a key characteristic of MoE-LLMs, we propose a compression method specifically designed for MoE-LLMs that combines static quantization and dynamic pruning to enhance their deployment efficiency. Our methods significantly reduce memory usage and improve inference speed while maintaining high model performance.

\section{Expert Pruning for MoE-LLMs}
Recently we have observed that expert pruning for MoE-LLMs has emerged as a prominent area of research. Based on this, we aim to offer a relatively comprehensive overview of related studies for the reference of other researchers.

First, we narrow the scope to post-training expert pruning, which can generally be divided into two main categories: static expert pruning and dynamic expert pruning. Below, we present the latest advancements under each category:

\noindent\textbf{(1) static expert pruning:} To the best of our knowledge, \citep{lu-etal-2024-experts} was the first to propose using a calibration set to determine the usage frequency of each expert in MoE and to evaluate expert importance based on their usage frequency. Experts with lower usage frequency are pruned prior to inference. Building on this, \citep{xie2024moeprunerpruningmixtureofexpertslarge} applied knowledge distillation after pre-inference expert pruning to restore the performance of the pruned MoE model. Similar to \citep{lu-etal-2024-experts} but not identical, \citep{DBLP:journals/corr/abs-2404-05089} leveraged the cumulative scores from the router's softmax output to assess expert importance for pre-inference pruning. Meanwhile, \citep{liu2024efficientexpertpruningsparse} introduced an expert merging approach, where less frequently used experts are merged with others to maintain the overall performance of the pruned MoE.

\noindent\textbf{(2) dynamic expert pruning:} Research on post-training dynamic expert pruning is relatively limited. \citep{lu-etal-2024-experts}
also proposed a dynamic pruning method based on the weight differences between experts selected for each token. Specifically, when the weight of the top-1 selected expert exceeds that of the top-2 expert by a certain threshold, only the top-1 expert is used. \citep{huang2024mc} further introduced a critical token protection mechanism in dynamic pruning.
\section*{Limitations}
Our method can significantly reduce memory consumption and improve inference speed while maintaining the performance of MoE models. However, there are still certain limitations to our approach:
(1) The proposed dynamic pruning method (PESF) calculates expert selection frequencies based on the current input sequence and determines the experts to prune accordingly. This method is only applicable during the prefill stage of model inference but is not suitable for the generate stage, where only a single token is input at a time. In the future, we aim to explore an MoE model pruning method that considers both inference phases, enabling inference acceleration benefits for the whole inference phase. \\
(2) We validated the effectiveness of our method on two MoE models with approximately 50B parameters and two MoE models with approximately 15B parameters. However, due to limited computational resources, we have not yet tested our method on larger-scale MoE models. For example, a recent significant breakthrough in the MoE field is the open release of DeepSeek-V3 \citep{deepseekai2024deepseekv3technicalreport}, which have a total of 671B parameters and 37B active parameters. Deepseek-V3 demonstrates comprehensive performance that even match or surpass some leading closed-source models\citep{openai2024gpt4technicalreport}. However, its enormous parameter count poses significant challenges for practical deployment. In the future, we will continue to explore quantization and pruning techniques for larger-scale MoE models, aiming to contribute to the advancement of MoE. \\
(3) The quantization method we propose, QESC, is theoretically orthogonal to other existing approaches that primarily focus on reducing quantization error itself, such as \citep{frantar-gptq, wang2018two, ashkboos2024quarot, wang2020towards, shao2023omniquant}. However, in our experiments, apart from GPTQ, we have not yet evaluated the performance of our QESC method when combined with these techniques. In future work, we hope to further investigate the overall effectiveness of integrating our quantization method with these established approaches. Additionally, similar to \citep{huang2024empirical, xiao2023robustmq}, we also aspire to validate our method across a broader range of benchmarks.

\section*{Ethics Statement}
In this work, we analyze the expert selection preferences of Mixture-of-Experts (MoE) models not only across three common types of NLP tasks—QA/CR, Math, and Code—but also including datasets in specific languages, with French being the selected language for the latter category.

We explicitly emphasize that the use of French in this study does not imply any bias, preference, or discriminatory intent towards or against any specific language, culture, or group of people. The selection is made solely to illustrate that MoE models exhibit different expert selection preferences across datasets in different languages.

\section*{Acknowledgments}
This work was supported in part by the National Key R\&D Program of China (No.2022ZD0160304), the Strategic Priority Research Program of Chinese Academy of Sciences(No.XDA0480200), the Beijing Municipal Science and Technology Project (No.Z231100010323002), and the Beijing Natural Science Foundation (No.L244046).

\bibliography{custom}

\appendix
\clearpage
\section{Appendix}
\label{sec:appendix}

\subsection{Time Consumption Analysis}
\noindent\textbf{QESC.} The quantization process of QESC primarily consists of two parts: GPTQ and the calibrating router, and can be executed on a single A100 40G GPU. We record the time spent on the GPTQ process and the calibration of routers separately, and calculate their respective proportions of the total time.
As shown in ~\Cref{tab:time-cost}, GPTQ accounts for the vast majority of the total quantization time, while the calibration of routers takes an average of only 1.94\% of the overall process. This demonstrates that our method introduces only a minimal additional time overhead to the quantization process, making it well-suited for practical applications.

\noindent\textbf{PESF.} Our PESF method introduces only a single-step online computation, as shown in ~\Cref{eq:pruning_formula}, resulting in virtually no additional delay.

\begin{table}[ht]
    \caption{Time consumption analysis for GPTQ and router calibration in our QESC method.
}
    \label{tab:time-cost}
    \centering
    \resizebox{0.48\textwidth}{!}{ 
    \begin{tabular}{c|c||cc}
        \hline
        \hline
        \multirow{1}{*}{\parbox{2cm}{\centering Models}} & \multirow{1}{*}{\centering Step of QESC} & Time(h) & Proportion(\%) \\
        \hline
        \multirow{2}{*}{Mixtral-8x7B} 
        & GPTQ & 1.30 & 98.48 \\
        ~ & Calibrating Router & 0.02 & 1.52 \\ 
        \hline
        \multirow{2}{*}{Phi3.5-moe} 
        & GPTQ & 1.39 & 97.89 \\ 
        ~ & Calibrating Router & 0.03 & 2.11 \\ 
        \hline
        \multirow{2}{*}{\makecell{Deepseek-moe\\-16b-base}} 
        & GPTQ & 1.75 & 97.22 \\ 
        ~ & Calibrating Router & 0.05 & 2.78 \\ 
        \hline
        \multirow{2}{*}{\makecell{Qwen1.5-MoE\\-A2.7B}} 
        & GPTQ & 1.48 & 98.67 \\ 
        ~ & Calibrating Router & 0.02 & 1.33 \\ 
        \hline
        \hline
    \end{tabular}
    }
    \vskip -0.2in
\end{table}

\subsection{Performance on Challenging Tasks}
\label{sec:challengeing_tasks}
In addition to validating the performance of our QESC method in maintaining model performance through Perplexity (PPL) and accuracies on common-sense intelligence tasks, we further evaluate our approach on more challenging mathematical task GSM8K \citep{cobbe2021gsm8k} and code generation task HumanEval \citep{chen2021evaluating}. We compare our method with three other methods—GPTQ, BSP, and PMQ. These two evaluations are conducted within the Bigcode-Evaluation-Harness \citep{bigcode-evaluation-harness} testing framework. 

As shown in \Cref{tab:challenging_performance}, under three different average quantization bit widths, our QESC method significantly outperforms the other three methods in preserving the performance of the post-quantization model on two challenging tasks. Notably, despite prior studies \citep{li2024evaluatingquantizedlargelanguage} indicating that post-quantization models are often highly sensitive to complex mathematical and coding tasks, we manage to limit the performance degradation of Mixtral-8x7B on GSM8K to within 3\% under 3.03-bit quantization, highlighting the effectiveness of our expert-selection calibration method.
\begin{table}[t]
    \caption{Comparison of the performance on challenging tasks GSM8K and HumanEval on Mixtral-8x7B. In the HumanEval evaluation, the hyperparameters are set as follows: temperature = 0.2, and nsamples = 10.
}
    \label{tab:challenging_performance}
    \centering
    \resizebox{0.48\textwidth}{!}{ 
    \begin{tabular}{c|c||ccc}
        \hline
        \multirow{1}{*}{\parbox{2cm}{\centering Models}} & \multirow{1}{*}{\parbox{2cm}{\centering Method}} & GSM8K & HumanEval (pass@10) \\
        \hline
        \multirow{1}{*}{16.00} & Full Precision & \num{58.30} & 59.15 \\ 
        \hline
        \multirow{3}{*}{2.06} 
        & GPTQ & 33.97 & 21.13 \\
        & PMQ & 20.17 & 11.91 \\
        ~ & \cellcolor{gray!15}\makebox[0pt][c]{\textbf{QESC}} & \cellcolor{gray!15}\makebox[0pt][c]{\textbf{37.15}} & \cellcolor{gray!15}\makebox[0pt][c]{\textbf{26.83}} \\ 
        \hline
        \multirow{4}{*}{2.54} 
        & GPTQ & 38.97 & 27.24 \\ 
        & BSP & \num{40.00} & 31.62 \\
        & PMQ & 36.94 & 29.77 \\
        ~ & \cellcolor{gray!15}\makebox[0pt][c]{\textbf{QESC}} & \cellcolor{gray!15}\makebox[0pt][c]{\textbf{43.14}} & \cellcolor{gray!15}\makebox[0pt][c]{\textbf{33.54}} \\ 
        \hline
        \multirow{3}{*}{3.03} 
        & GPTQ & 52.29 & 40.83 \\ 
        & BSP & 53.72 & 42.07 \\
        ~ & \cellcolor{gray!15}\makebox[0pt][c]{\textbf{QESC}} & \cellcolor{gray!15}\makebox[0pt][c]{\textbf{55.34}} & \cellcolor{gray!15}\makebox[0pt][c]{\textbf{46.34}} \\ 
        \hline
        \hline
    \end{tabular}
    }
    \vskip -0.2in
\end{table}

\subsection{Overfitting Analysis of Mixed-Precision Quantization Methods for MoE Models}
\label{sec:overfitting}
In ~\Cref{sec:generalization}, we observe that the importance of the same expert may vary drastically across different tasks. Based on this insight, we deduce that using any calibration set to determine the importance of an expert before inference may be biased and lack generalization. Here, we further substantiate this point through comparative experiments. Specifically, we first utilize QA/CR datasets, Math datasets, Code datasets, datasets in French version, and the C4 dataset \citep{raffel2020exploring} as calibration sets, obtaining five distinct expert selection frequencies from these calibration sets. Subsequently, we automatically allocate the quantization bit-width for each expert based on the expert selection frequencies, following the algorithm mentioned in the state-of-the-art mixed-precision quantization method for MoE, PMQ \citep{huang2024mc} (details is shown in ~\Cref{sec:repoduce-quant}). We then quantize the model to an average bit-width of 2.06 bits using the five quantization bit-width settings derived (the calibration set used in the paper is the C4 dataset). Finally, we evaluate the performance of these five quantized models on four datasets: Hellswag (QA/CR), MathQA (Math), Lambada\_fr (French), and Conala (Code), each representing a different task category. Additionally, we compare the performance of our QESC method at the same average quantization bit-width.

\begin{table*}[ht]
    \caption{A comparison of the performance of quantized Mixtral-8x7B and Deepseek-moe-16b-base, based on the mixed-precision quantization method PMQ, using five different calibration sets under an average quantization bit-width of 2.06 bits, across four types of task datasets.}
    \label{tab:overfitting}
    \centering
    \resizebox{\textwidth}{!}{ 
    \begin{tabular}{c|c|c|c||cccc}
        \hline
        \hline
         Models & Bits & Method & Calibration Dataset & \makecell{Hellaswag \\ (QA/CR)} & \makecell{MathQA \\ (Math)} & \makecell{Lambada\_fr \\ (French)} & \makecell{Conala \\ (Code)} \\
        \hline
        \multirow{7}{*}{\centering Mixtral-8x7B} 
        & 16.00 & Baseline & None & 84.03 & 41.64 & 65.96 & 73.86 \\
        \cline{2-8}
        ~ & \multirow{5}{*}{2.06} & \multirow{5}{*}{\makecell{Mixed-\\Precision}} & QA/CR & \cellcolor{gray!20}\makebox[0pt][c]{75.49} & 31.06 & 49.17 & 7.25 \\
        ~ & ~ & ~ & Math & 69.25 & \cellcolor{gray!15}\makebox[0pt][c]{32.26} & 47.06 & 7.64 \\
        ~ & ~ & ~ & French & 65.55 & 24.39 & \cellcolor{gray!15}\makebox[0pt][c]{51.87} & \num{3.80} \\
        ~ & ~ & ~ & Code & 67.67 & 31.66 & 44.50 & \cellcolor{gray!15}\makebox[0pt][c]{37.42} \\
        ~ & ~ & ~ & C4 & 74.95 & 31.79 & 49.12 & 16.22 \\
        \cline{2-8}
        ~ & 2.06 & \textbf{QESC} & None & \textbf{77.27} & \textbf{36.08} & \textbf{60.37} & \textbf{66.68} \\ 
        \hline
        \multirow{7}{*}{\centering \makecell{Deepseek-moe \\ -16b-base}} 
        & 16.00 & Baseline & None & 77.43 & 31.66 & 58.08 & 58.76 \\ 
        \cline{2-8}
        ~ & \multirow{5}{*}{2.06} & \multirow{5}{*}{\makecell{Mixed-\\Precision}} & QA/CR & \cellcolor{gray!15}\makebox[0pt][c]{72.09} & 24.25 & 45.76 & 7.57 \\
        ~ & ~ & ~ & Math & 62.15 & \cellcolor{gray!15}\makebox[0pt][c]{30.52} & 41.74 & 8.69 \\
        ~ & ~ & ~ & French & 65.49 & 23.15 & \cellcolor{gray!15}\makebox[0pt][c]{48.86} & \num{1.80} \\
        ~ & ~ & ~ & Code & 54.87 & 29.61 & 35.88 & \cellcolor{gray!15}\makebox[0pt][c]{26.34} \\
        ~ & ~ & ~ & C4 & 68.70 & 26.26 & 42.13 & 16.22 \\
        \cline{2-8}
        ~ & 2.06 & \textbf{QESC} & None & \textbf{68.85} & \textbf{28.91} & \textbf{52.38} & \textbf{42.36} \\ 
        \hline
        \hline
    \end{tabular}
    }
\end{table*}

As shown in ~\Cref{tab:overfitting}, the mixed-precision quantization method based on expert usage frequency exhibits significant overfitting on both models. As highlighted by the red-marked values in the table, using a specific task dataset as the calibration set results in a post-quantization model that achieves relatively optimal performance on that specific task but shows severe performance degradation on other tasks. This overfitting phenomenon is most pronounced in Code-related tasks. When calibration is performed using datasets from other task types, the quantized models experience a drastic performance collapse on Code tasks. Only the models calibrated with Code-specific task datasets manage to achieve relatively good results on the Conala dataset.

When the C4 dataset is used as the calibration set, due to its inherently balanced nature, the quantized models achieve relatively average performance across all four task types. However, even in this case, there is still severe performance degradation on Code-related tasks. In contrast, our proposed QESC method focuses on the calibration of expert-selection. It significantly outperforms mixed-precision method across all four datasets, demonstrating superior generalization capabilities.

\subsection{Search of Optimal Value of K in TopK-MSE Loss}
\label{sec:best_k_in_topk}
In this subsection, we explore the optimal k-value settings for TopK-MSE in three MoE models with relatively larger numbers of total experts: Phi3.5-moe, Deepseek-moe-16b-base, and Qwen1.5-moe-A2.7B, under three average quantization bit widths—2.06 bits, 2.56 bits, and 3.03 bits. For Phi3.5-moe (selecting 2 experts out of 16), we set the k-values to 4, 6, 8, 10, 12, and 16 (equivalent to MSE loss). For Deepseek-moe-16b-base (selecting 6 experts out of 64) and Qwen1.5-moe-A2.7B (selecting 4 experts out of 64), the k-values are set to 8, 12, 16, 20, 24, 32, 48, and 64 (equivalent to MSE loss). For each k-value, we quantize the MoE models to the three average bit widths while keeping other configurations constant and compare their performance on the MMLU dataset \citep{hendrycks2021ethics}, which serves as a comprehensive benchmark to effectively evaluate the models' capabilities across diverse tasks.

\begin{figure*}[htb] 
    \centering
    \begin{subfigure}[b]{0.32\linewidth} 
        \centering
        \includegraphics[width=1.0\linewidth]{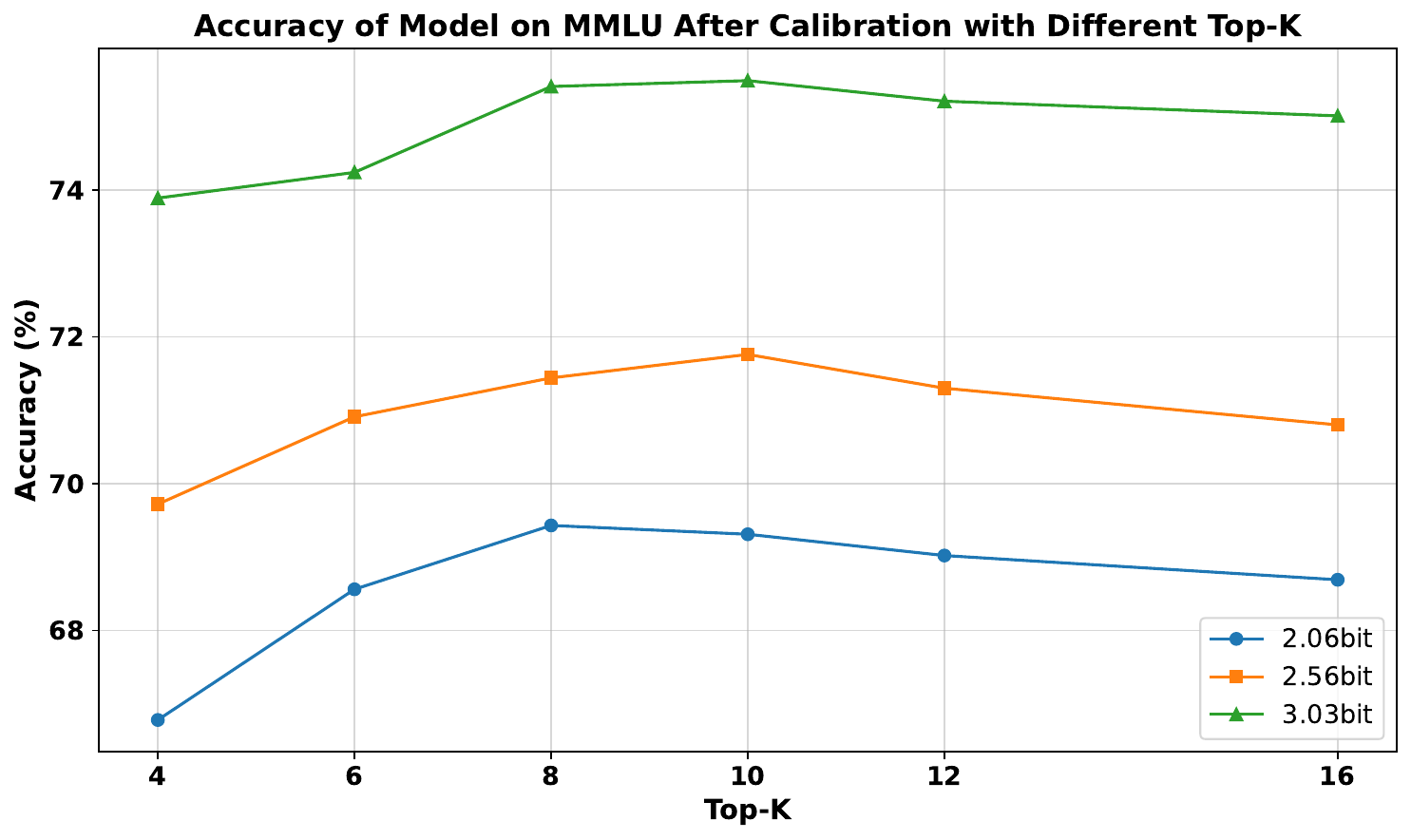} 
        \caption{Phi3.5-moe}
        \label{fig:topk-phimoe}
    \end{subfigure}
    \hfill 
    \begin{subfigure}[b]{0.32\linewidth} 
        \centering
        \includegraphics[width=1.0\linewidth]{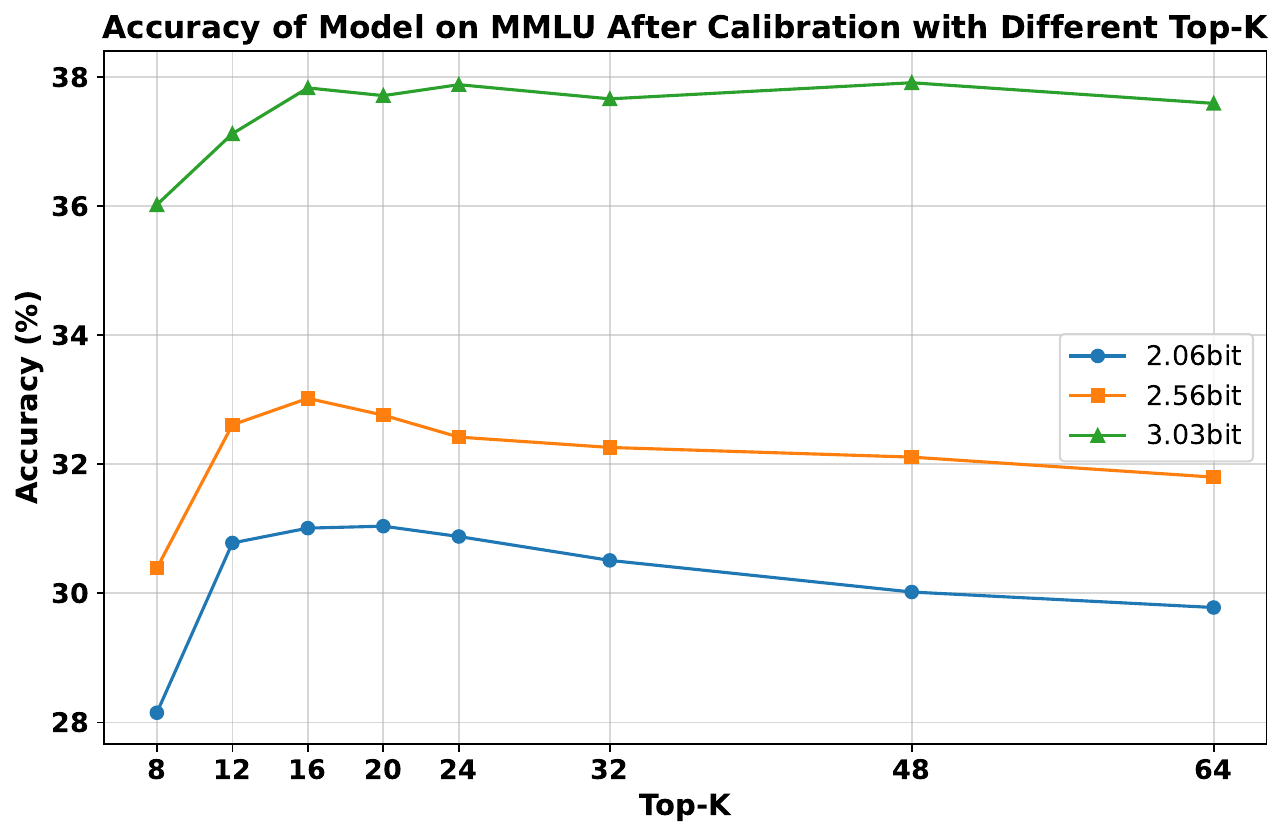} 
        \caption{Deepseek-moe-16b-base}
        \label{fig:topk-deepseek}
    \end{subfigure}
    \hfill 
    \begin{subfigure}[b]{0.32\linewidth} 
        \centering
        \includegraphics[width=1.0\linewidth]{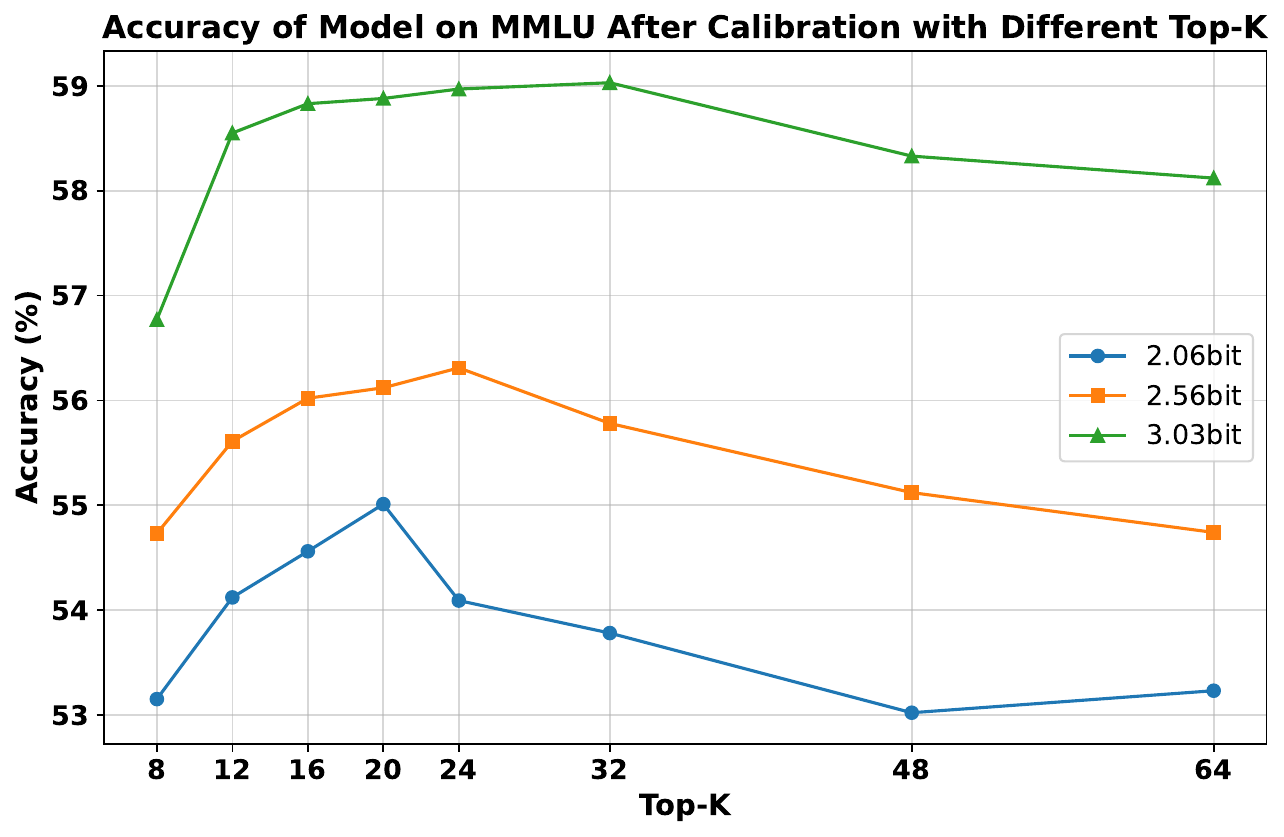} 
        \caption{Qwen1.5-MoE-A2.7B}
        \label{fig:topk-qwen}
    \end{subfigure}

    \caption{The accuracy of Phi3.5-moe (left), Deepseek-moe-16b-base (middle), and Qwen1.5-MoE-A2.7B (right) on the MMLU dataset varies under different TopK-MSE optimization under 2.06bit, 2.56bit and 3.03bit quantization.}
    \label{fig:topk-search}
\end{figure*}

As shown in ~\Cref{fig:topk-search}, when the k-value is too small and approaches the number of experts selected per token in the model, significant performance degradation occurs. Referring to ~\Cref{fig:topk_plot}, this may be attributed to a certain degree of overfitting. For Phi3.5-moe, the optimization results with k-values of 8–10 under all three average quantization bit widths are noticeably better than those with a k-value of 16. For Deepseek-moe-16b-base and Qwen1.5-moe-A2.7B, the optimal results are observed at k-values of 16–24 under 2.06-bit and 2.56-bit quantization. This aligns with our observations in ~\Cref{fig:topk_plot}, where our optimization effectively covers over 95\% of the incorrectly unselected experts while avoiding the noise introduced by experts with low selection probabilities. Under the 3.03-bit quantization, however, the optimization results show minimal variation with changes in k-value, as the higher quantization bit width inherently results in a lower rate of expert selection changes.

Based on the above observations, as shown in ~\Cref{tab:topk-tab}, we set the k-values to 8, 20, and 20 for expert selection calibration for the three models, respectively. All quantization-related experimental results in this work are based on this configuration.

\begin{table}[htbp]
\centering
\caption{K Values for expert-selection in MoE models using TopK-MSE calibration}
\label{tab:topk-tab}
\resizebox{0.48\textwidth}{!}{%
\begin{tabular}{lccc}
\toprule
\textbf{Model} & \textbf{Total Experts} & \textbf{Experts per Token} & \textbf{K} \\
\midrule
Phi3.5-moe            & 16 & 2 & 8 \\
Deepseek-moe-16b-base & 64 & 6 & 20 \\
Qwen1.5-MoE-A2.7B     & 64 & 4 & 20 \\
\bottomrule
\end{tabular}%
}
\end{table}

\subsection{Quantization Bit-Width Settings}
\label{sec:detailed-bit}

\begin{table}[bp]
\centering
\caption{Proportion of non-embedding parameters in MoE models}
\label{tab:params_analysis}
\resizebox{0.48\textwidth}{!}{%
\begin{tabular}{lccc}
\toprule
\textbf{Model} & \textbf{MHSA (\%)} & \textbf{Experts (\%)} & \textbf{Router (\%)} \\
\midrule
Mixtral-8x7B          & 2.894\% & 97.104\% & 0.002\% \\
Phi3.5-moe            & 3.226\% & 96.774\% & 0.005\% \\
Deepseek-moe-16b-base & 2.852\% & 97.122\% & 0.022\% \\
Qwen1.5-MoE-A2.7B     & 2.945\% & 97.039\% & 0.022\% \\
\bottomrule
\end{tabular}%
}
\end{table}

\begin{table}[bp]
\centering
\caption{The average quantization bit-width of the four models when the experts' quantization bit-width is set to 2-bit, 2.5-bit, and 3-bit.}
\label{tab:avg_bits}
\resizebox{0.48\textwidth}{!}{%
\begin{tabular}{lccc}
\toprule
\textbf{Model} & \textbf{2-bit} & \textbf{2.5-bit} & \textbf{3-bit} \\
\midrule
Mixtral-8x7B           & 2.058 & 2.544 & 3.029 \\
Phi3.5-moe             & 2.065 & 2.549 & 3.033 \\
Deepseek-moe-16b-base  & 2.060 & 2.546 & 3.031 \\
Qwen1.5-MoE-A2.7B      & 2.062 & 2.547 & 3.032 \\
\bottomrule
\end{tabular}%
}
\end{table}

\begin{figure*}[htb] 
    \centering
    \begin{subfigure}[b]{0.48\linewidth} 
        \centering
        \includegraphics[width=1.0\linewidth]{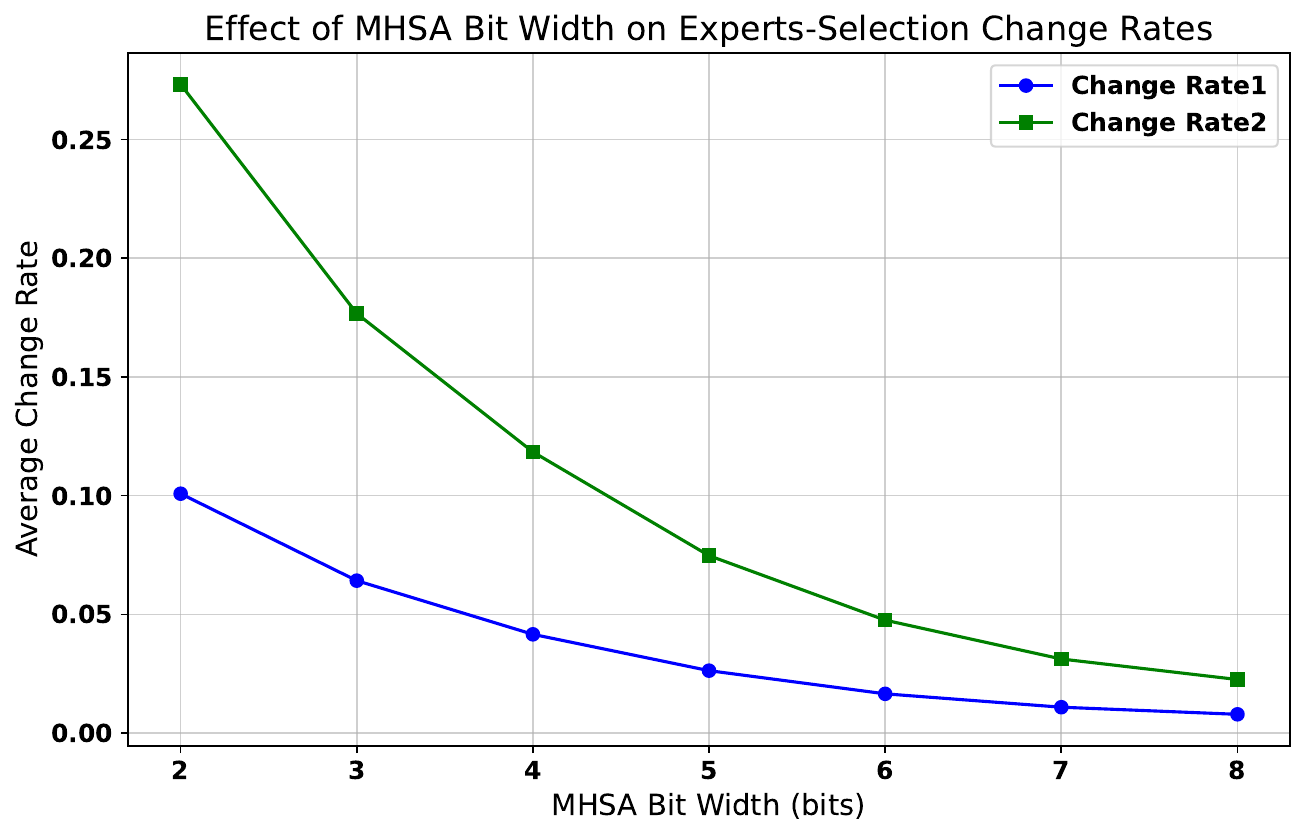} 
        \label{fig:attn_freq}
    \end{subfigure}
    \hfill 
    \begin{subfigure}[b]{0.48\linewidth} 
        \centering
        \includegraphics[width=1.0\linewidth]{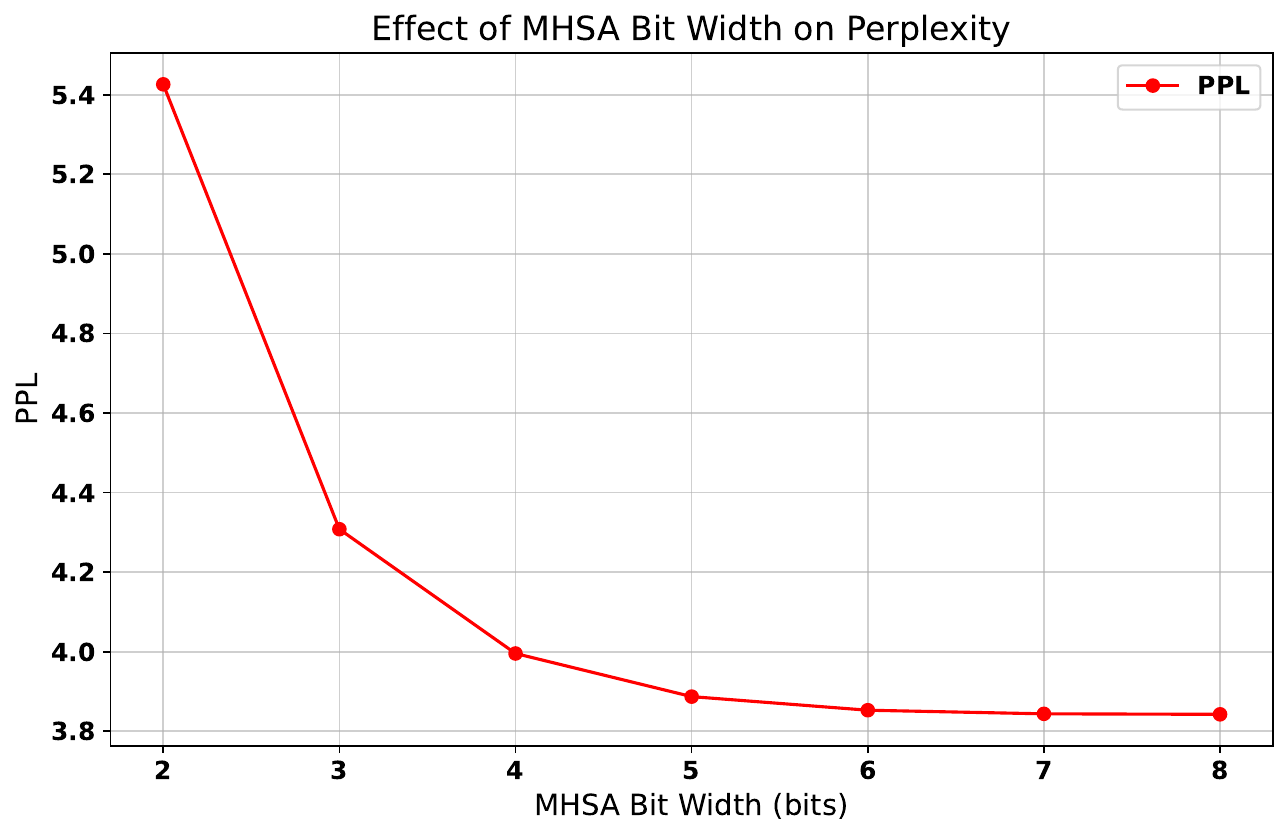} 
        \label{fig:attn_ppl}
    \end{subfigure}

    \caption{The changes in Experts-Selection rates and WikiText2 perplexity (PPL) with varying MHSA quantization bit-widths on Mixtral-8x7B. "Change Rate 1" corresponds to cases where both selected experts are changed, while "Change Rate 2" corresponds to cases where one or more expert selections are altered.}
    \label{fig:MHSA}
\end{figure*}

In MoE-LLMs, the vast majority of non-embedding parameters come from the experts within the MoE layers, while the router in the MoE layer and MHSA account for only about 3\% of the total parameters. For the four MoE models used in the experiments of this paper—Mixtral-8x7B, Deepseek-moe-16b-base, Phi3.5-moe, and Qwen1.5-MoE-A2.7B—we first conduct a detailed analysis of the parameter proportions in each component, as shown in ~\Cref{tab:params_analysis}. Considering that the router accounts for less than 0.03\% of the total parameters, yet plays a crucial role in expert selection, this work retains the router at its original precision.

Unlike the experts in the MoE layer, the MHSA component affects all tokens in the input sequence. Previous studies \citep{li2024examining} have compared the impact of increasing the quantization bit-width of MHSA and the MoE layer on overall model performance, concluding that MHSA is more bit-efficient. In this work, we investigate the effect of MHSA quantization bit-width on overall model performance from the perspective of expert selection change rate. Using Mixtral-8x7B, we conduct the following experiments: keeping the rest of the model at its original precision, the MHSA component is quantized to bit-widths ranging from 2-bit to 8-bit. We then calculate the average expert selection change rate relative to the original model and the perplexity (PPL) of the quantized model on the WikiText2 \citep{merity2016pointer}. 

As shown in ~\Cref{fig:MHSA}, within the 2-4 bit range, both the expert selection change rate and PPL are highly sensitive to the quantization bit-width. As the bit-width increases, the expert selection change rate and PPL decrease significantly. This demonstrates that maintaining MHSA at a relatively high quantization bit-width is indeed important for ensuring the quantized model can still select the correct experts and maintain overall model performance. Considering that in the 4-8 bit range, both metrics change more gradually with increasing bit-width, and from a hardware perspective, current systems are unable to achieve efficient acceleration for bit-widths like 5-bit or 6-bit, we choose to set the MHSA quantization bit-width to 4-bit. This choice strikes a balance between maintaining the performance of the quantized model and minimizing overall memory usage.

By comprehensively analyzing the above discussion, we quantize the MHSA section to 4-bit, retaine the router at its original precision, and quantize the experts to 2-bit, 2.5-bit, and 3-bit. Under the 2.5-bit setting, we align with previous research \citep{li2024examining}, which finds that the earlier layers in MoE benefit from higher quantization bit-widths. Therefore, we simply quantize the experts in the first half of the layers to 3-bit and those in the second half to 2-bit.
Based on this configuration, we calculate the average bit-width of the experts for the four models under the 2-bit, 2.5-bit, and 3-bit conditions, as shown in ~\Cref{tab:avg_bits}. Since the parameter proportions of different components vary slightly across the four models, the final average bit-width also exhibits minor differences. However, as these differences are negligible, for simplicity, we represent the average quantization bit-widths of the four models in this paper as 2.06-bit, 2.54-bit, and 3.03-bit for the three respective settings.

\subsection{Reproduction Details of Quantization}
\label{sec:repoduce-quant}
BSP \citep{li2024examining} and PMQ \citep{huang2024mc} are two methods built upon GPTQ that focus on mixed-precision quantization. BSP reports its results on Mixtral-8x7B and Deepseek-moe-16b-base in its paper, while PMQ only reports results on Mixtral-8x7B. To ensure a fair comparison, we refer to both the papers and official repositories of these methods, and apply their respective quantization bit-width allocation strategies to the models used in this study. We then perform GPTQ quantization and evaluate the final results using the same framework as this paper.

Specifically, for BSP, we first use the same calibration datasets mentioned in its paper (WikiText2 dataset) to obtain the expert usage frequencies for Phi3.5-moe and Qwen1.5-MoE-A2.7B. In Phi3.5-moe, each MoE layer selects 2 experts out of 16. Following BSP's settings for Mixtral-8x7B, at the 3.03-bit bit-width configuration, we allocate 4-bit to the top-8 most frequently used experts and 2-bit to the remaining 8 experts. At the 2.54-bit configuration, we allocate 3-bit to the top-8 experts and 2-bit to the remaining 8 experts. For Qwen1.5-MoE-A2.7B, each MoE layer selects 4 experts out of 60, with an additional 4 shared experts. Following BSP's settings for Deepseek-moe-16b-base, all shared experts are allocated 8-bit. At the 3.03-bit configuration, we allocate 4-bit to the top-20 most frequently used experts and 2-bit to the remaining 40 experts. At the 2.54-bit configuration, we allocate 4-bit to the top-6 experts and 2-bit to the remaining 54 experts.

For PMQ, we also use the same calibration datasets mentioned in its paper (C4 dataset \citep{raffel2020exploring}) to obtain expert usage frequencies for all four models. For Mixtral-8x7B and Phi3.5-moe, we directly apply the Integer Programming (IP) optimization algorithm used in PMQ's paper to derive the corresponding mixed-precision bit-width configurations. For Deepseek-moe-16b-base and Qwen1.5-MoE-A2.7B, as PMQ's paper did not discuss MoE models with shared experts, we use the same IP optimization algorithm to determine the mixed-precision bit-width configurations for non-shared experts under the average bit-width settings of 2.06-bit and 2.54-bit. For shared experts, we allocate 2-bit at the 2.06-bit configuration and 3-bit at the 2.54-bit configuration to ensure fairness in comparative experiments with PMQ.

\subsection{Completed Results of Quantization}
\label{sec:all-results}
In ~\Cref{tab:quant-mixtral-8x7B,tab:quant-phi3.5-moe,tab:quant-deepseek-moe-16b-base,tab:quant-qwen1.5-moe-a2.7b}, we present the full results of ~\Cref{tab:quantization_comparison}, including detailed accuracies on eight zero-shot tasks, and compare our approach with GPTQ \citep{frantar-gptq}, BSP \citep{li2024examining} and PMQ \citep{huang2024mc}.

\begin{table*}[!ht]
    \centering
    \caption{Comprehensive comparison of average accuracy of Mixtral-8x7B on 8 zero-shot tasks under quantization.}
    \label{tab:quant-mixtral-8x7B}
    \resizebox{\textwidth}{!}{
    \begin{tabular}{c|c|ccccccccccc}
    \hline
    \hline
        \parbox{2cm}{\centering Bits} & Method & PIQA & ARC-E & ARC-C & BOOLQ & HS & WG & MATHQA & MMLU & Avg \\ 
        \hline
        \multirow{1}{*}{16.00} & Full Precision & 83.57 & 83.67 & 59.3 & 85.35 & 84.03 & 76.24 & 41.64 & 67.29 & 72.64 \\ 
        \hline
        \multirow{3}{*}{2.06} & GPTQ & 76.16 & 75.47 & 47.91 & \num{77.40} & 71.57 & 71.98 & 31.84 & 48.14 & 62.56 \\ 
        ~ & PMQ & 79.16 & 73.06 & 48.38 & 80.58 & 74.95 & 71.27 & 31.79 & \num{46.80} & 63.25 \\ 
        ~ & \cellcolor{gray!15}\makebox[0pt][c]{EAC-MoE} & \cellcolor{gray!15}\makebox[0pt][c]{79.16} & \cellcolor{gray!15}\makebox[0pt][c]{\num{76.30}} & \cellcolor{gray!15}\makebox[0pt][c]{51.02} & \cellcolor{gray!15}\makebox[0pt][c]{76.79} & \cellcolor{gray!15}\makebox[0pt][c]{77.27} & \cellcolor{gray!15}\makebox[0pt][c]{74.82} & \cellcolor{gray!15}\makebox[0pt][c]{36.08} & \cellcolor{gray!15}\makebox[0pt][c]{59.01} & \cellcolor{gray!15}\makebox[0pt][c]{66.31} \\ 
        \hline
        \multirow{4}{*}{2.54} & GPTQ & 80.56 & 79.04 & 56.06 & 85.84 & \num{78.20} & 73.01 & 37.62 & 58.89 & 68.65 \\ 
        ~ & BSP & 80.96 & 77.86 & 53.24 & 72.53 & 77.95 & 73.56 & 35.34 & 52.04 & 65.44 \\ 
        ~ & PMQ & 80.52 & \num{77.10} & 51.28 & 82.54 & 79.03 & 73.95 & 39.18 & 56.37 & \num{67.50} \\ 
        ~ & \cellcolor{gray!15}\makebox[0pt][c]{EAC-MoE} & \cellcolor{gray!15}\makebox[0pt][c]{81.45} & \cellcolor{gray!15}\makebox[0pt][c]{80.77} & \cellcolor{gray!15}\makebox[0pt][c]{55.03} & \cellcolor{gray!15}\makebox[0pt][c]{84.85} & \cellcolor{gray!15}\makebox[0pt][c]{79.44} & \cellcolor{gray!15}\makebox[0pt][c]{\num{75.3}} & \cellcolor{gray!15}\makebox[0pt][c]{37.86} & \cellcolor{gray!15}\makebox[0pt][c]{61.37} & \cellcolor{gray!15}\makebox[0pt][c]{69.51} \\ 
        \hline
        \multirow{3}{*}{3.03} & GPTQ & 80.63 & 79.21 & 55.55 & 84.59 & 79.63 & 74.43 & 37.49 & 59.82 & 68.92 \\ 
        ~ & BSP & 81.56 & 79.71 & 54.59 & 75.75 & 80.06 & 74.74 & 36.68 & 54.69 & 67.22 \\ 
        ~ & \cellcolor{gray!15}\makebox[0pt][c]{EAC-MoE} & \cellcolor{gray!15}\makebox[0pt][c]{82.97} & \cellcolor{gray!15}\makebox[0pt][c]{83.25} & \cellcolor{gray!15}\makebox[0pt][c]{58.87} & \cellcolor{gray!15}\makebox[0pt][c]{85.47} & \cellcolor{gray!15}\makebox[0pt][c]{82.77} & \cellcolor{gray!15}\makebox[0pt][c]{\num{76.40}} & \cellcolor{gray!15}\makebox[0pt][c]{41.81} & \cellcolor{gray!15}\makebox[0pt][c]{\num{66.10}} & \cellcolor{gray!15}\makebox[0pt][c]{72.21} \\ 
        \hline
        \hline
    \end{tabular}}
\end{table*}

\begin{table*}[!ht]
    \centering
    \caption{Comprehensive comparison of average accuracy of Phi3.5-moe on 8 zero-shot tasks under quantization.}
    \label{tab:quant-phi3.5-moe}
    \resizebox{\textwidth}{!}{
    \begin{tabular}{c|c|ccccccccccc}
    \hline
    \hline
        \parbox{2cm}{\centering Bits} & Method & PIQA & ARC-E & ARC-C & BOOLQ & HS & WG & MATHQA & MMLU & Avg \\ 
        \hline
        \multirow{1}{*}{16.00} & Full Precision & 77.69 & \num{65.70} & 53.58 & 88.35 & 79.87 & \num{76.80} & 38.32 & 76.62 & 69.62 \\ 
        \hline
        \multirow{3}{*}{2.06} & GPTQ & 75.57 & 57.87 & 48.46 & 86.88 & 73.01 & 71.72 & \num{32.80} & 69.29 & 64.45 \\ 
        ~ & PMQ & 75.68 & 50.59 & 41.55 & 86.18 & 78.45 & 73.86 & 22.31 & 62.14 & 61.35 \\ 
        ~ & \cellcolor{gray!15}\makebox[0pt][c]{EAC-MoE} & \cellcolor{gray!15}\makebox[0pt][c]{76.17} & \cellcolor{gray!15}\makebox[0pt][c]{59.47} & \cellcolor{gray!15}\makebox[0pt][c]{49.06} & \cellcolor{gray!15}\makebox[0pt][c]{\num{86.80}} & \cellcolor{gray!15}\makebox[0pt][c]{73.34} & \cellcolor{gray!15}\makebox[0pt][c]{71.35} & \cellcolor{gray!15}\makebox[0pt][c]{34.64} & \cellcolor{gray!15}\makebox[0pt][c]{69.43} & \cellcolor{gray!15}\makebox[0pt][c]{65.03} \\ 
        \hline
        \multirow{4}{*}{2.54} & GPTQ & 77.37 & 61.49 & 48.98 & 86.67 & 75.32 & 72.61 & 32.86 & 71.19 & 65.81 \\ 
        ~ & BSP & 76.99 & 61.41 & 50.71 & 87.22 & 76.09 & 73.51 & 32.23 & 71.05 & 66.15 \\ 
        ~ & PMQ & 77.04 & 57.32 & 48.21 & 87.68 & 77.42 & 74.27 & 33.67 & 72.66 & 66.03 \\ 
        ~ & \cellcolor{gray!15}\makebox[0pt][c]{EAC-MoE} & \cellcolor{gray!15}\makebox[0pt][c]{77.09} & \cellcolor{gray!15}\makebox[0pt][c]{61.78} & \cellcolor{gray!15}\makebox[0pt][c]{51.19} & \cellcolor{gray!15}\makebox[0pt][c]{86.57} & \cellcolor{gray!15}\makebox[0pt][c]{75.56} & \cellcolor{gray!15}\makebox[0pt][c]{73.32} & \cellcolor{gray!15}\makebox[0pt][c]{35.08} & \cellcolor{gray!15}\makebox[0pt][c]{71.67} & \cellcolor{gray!15}\makebox[0pt][c]{66.53} \\ 
        \hline
        \multirow{3}{*}{3.03} & GPTQ & \num{78.10} & 63.17 & 51.96 & \num{87.70} & 78.81 & 74.59 & 35.64 & 75.01 & 68.12 \\ 
        ~ & BSP & \num{77.20} & \num{64.60} & 51.71 & 87.49 & 78.73 & 76.01 & 33.23 & \num{72.40} & 67.67 \\ 
        ~ & \cellcolor{gray!15}\makebox[0pt][c]{EAC-MoE} & \cellcolor{gray!15}\makebox[0pt][c]{78.24} & \cellcolor{gray!15}\makebox[0pt][c]{62.21} & \cellcolor{gray!15}\makebox[0pt][c]{52.56} & \cellcolor{gray!15}\makebox[0pt][c]{87.83} & \cellcolor{gray!15}\makebox[0pt][c]{78.59} & \cellcolor{gray!15}\makebox[0pt][c]{76.16} & \cellcolor{gray!15}\makebox[0pt][c]{36.85} & \cellcolor{gray!15}\makebox[0pt][c]{75.49} & \cellcolor{gray!15}\makebox[0pt][c]{68.49} \\ 
        \hline
        \hline
    \end{tabular}}
\end{table*}

\begin{table*}[!ht]
    \centering
    \caption{Comprehensive comparison of average accuracy of Deepseek-moe-16b-base on 8 zero-shot tasks under quantization.}
    \label{tab:quant-deepseek-moe-16b-base}
    \resizebox{\textwidth}{!}{
    \begin{tabular}{c|c|ccccccccccc}
    \hline
    \hline
        \parbox{2cm}{\centering Bits} & Method & PIQA & ARC-E & ARC-C & BOOLQ & HS & WG & MATHQA & MMLU & Avg \\ 
        \hline
        \multirow{1}{*}{16.00} & Full Precision & 80.52 & 73.19 & 47.53 & 72.57 & 77.43 & 69.93 & 31.66 & 38.18 & 61.38 \\ 
        \hline
        \multirow{3}{*}{2.06} & GPTQ & 76.77 & 65.78 & 40.02 & 67.34 & 69.41 & 65.43 & \num{27.50} & 26.76 & 54.88 \\ 
        ~ & PMQ & 77.64 & 64.73 & 39.85 & 67.75 & 68.7 & 66.22 & 26.26 & 27.16 & 54.79 \\ 
        ~ & \cellcolor{gray!15}\makebox[0pt][c]{EAC-MoE} & \cellcolor{gray!15}\makebox[0pt][c]{77.48} & \cellcolor{gray!15}\makebox[0pt][c]{70.24} & \cellcolor{gray!15}\makebox[0pt][c]{42.66} & \cellcolor{gray!15}\makebox[0pt][c]{\num{70.20}} & \cellcolor{gray!15}\makebox[0pt][c]{68.85} & \cellcolor{gray!15}\makebox[0pt][c]{67.01} & \cellcolor{gray!15}\makebox[0pt][c]{28.91} & \cellcolor{gray!15}\makebox[0pt][c]{31.04} & \cellcolor{gray!15}\makebox[0pt][c]{57.05} \\ 
        \hline
        \multirow{4}{*}{2.54} & GPTQ & 78.29 & 68.48 & 41.72 & 69.88 & 70.97 & 67.48 & 28.17 & 29.65 & 56.83 \\ 
        ~ & BSP & 78.89 & 70.24 & 42.32 & 74.34 & 72.96 & 68.59 & 27.87 & 30.68 & 58.24 \\ 
        ~ & PMQ & 78.78 & \num{69.70} & 41.89 & 74.29 & 71.27 & 67.96 & 29.25 & 32.89 & \num{58.00} \\ 
        ~ & \cellcolor{gray!15}\makebox[0pt][c]{EAC-MoE} & \cellcolor{gray!15}\makebox[0pt][c]{78.51} & \cellcolor{gray!15}\makebox[0pt][c]{69.11} & \cellcolor{gray!15}\makebox[0pt][c]{42.32} & \cellcolor{gray!15}\makebox[0pt][c]{74.59} & \cellcolor{gray!15}\makebox[0pt][c]{73.97} & \cellcolor{gray!15}\makebox[0pt][c]{69.14} & \cellcolor{gray!15}\makebox[0pt][c]{28.98} & \cellcolor{gray!15}\makebox[0pt][c]{33.02} & \cellcolor{gray!15}\makebox[0pt][c]{58.71} \\ 
        \hline
        \multirow{3}{*}{3.03} & GPTQ & 78.89 & 72.22 & 44.37 & 72.42 & 75.15 & 67.01 & 29.88 & 34.72 & 59.33 \\ 
        ~ & BSP & 79.68 & 72.39 & 44.37 & 73.98 & 74.61 & \num{68.90} & 28.14 & \num{33.10} & \num{59.40} \\ 
        ~ & \cellcolor{gray!15}\makebox[0pt][c]{EAC-MoE} & \cellcolor{gray!15}\makebox[0pt][c]{80.03} & \cellcolor{gray!15}\makebox[0pt][c]{73.15} & \cellcolor{gray!15}\makebox[0pt][c]{\num{45.90}} & \cellcolor{gray!15}\makebox[0pt][c]{75.29} & \cellcolor{gray!15}\makebox[0pt][c]{75.68} & \cellcolor{gray!15}\makebox[0pt][c]{70.48} & \cellcolor{gray!15}\makebox[0pt][c]{31.42} & \cellcolor{gray!15}\makebox[0pt][c]{37.83} & \cellcolor{gray!15}\makebox[0pt][c]{61.22} \\ 
        \hline
        \hline
    \end{tabular}}
\end{table*}

\begin{table*}[!ht]
    \centering
    \caption{Comprehensive comparison of average accuracy of Qwen1.5-MoE-A2.7B on 8 zero-shot tasks under quantization.}
    \label{tab:quant-qwen1.5-moe-a2.7b}
    \resizebox{\textwidth}{!}{
    \begin{tabular}{c|c|ccccccccccc}
    \hline
    \hline
        \parbox{2cm}{\centering Bits} & Method & PIQA & ARC-E & ARC-C & BOOLQ & HS & WG & MATHQA & MMLU & Avg \\ 
        \hline
        \multirow{1}{*}{16.00} & Full Precision & 80.79 & 69.44 & 44.37 & 79.57 & 77.17 & 69.77 & 35.57 & 61.08 & 64.72 \\ 
        \hline
        \multirow{3}{*}{2.06} & GPTQ & 75.79 & 65.53 & 40.02 & 72.14 & 67.06 & 64.48 & 30.02 & 47.07 & 57.76 \\ 
        ~ & PMQ & 76.14 & 65.69 & 40.11 & 69.88 & 68.71 & 64.52 & 29.01 & 48.23 & 57.79 \\ 
        ~ & \cellcolor{gray!15}\makebox[0pt][c]{EAC-MoE} & \cellcolor{gray!15}\makebox[0pt][c]{\num{78.40}} & \cellcolor{gray!15}\makebox[0pt][c]{65.28} & \cellcolor{gray!15}\makebox[0pt][c]{40.78} & \cellcolor{gray!15}\makebox[0pt][c]{70.37} & \cellcolor{gray!15}\makebox[0pt][c]{\num{71.50}} & \cellcolor{gray!15}\makebox[0pt][c]{65.98} & \cellcolor{gray!15}\makebox[0pt][c]{29.28} & \cellcolor{gray!15}\makebox[0pt][c]{54.56} & \cellcolor{gray!15}\makebox[0pt][c]{59.52} \\ 
        \hline
        \multirow{4}{*}{2.54} & GPTQ & 77.48 & 63.17 & 39.16 & 68.69 & 70.42 & 64.17 & 29.28 & 50.88 & 57.91 \\ 
        ~ & BSP & 79.54 & 65.03 & 39.59 & 68.99 & 73.77 & 68.51 & 31.89 & 55.91 & \num{60.40} \\ 
        ~ & PMQ & 78.16 & 65.75 & 41.21 & 71.34 & 72.34 & 68.01 & 31.66 & 55.31 & 60.47 \\ 
        ~ & \cellcolor{gray!15}\makebox[0pt][c]{EAC-MoE} & \cellcolor{gray!15}\makebox[0pt][c]{78.73} & \cellcolor{gray!15}\makebox[0pt][c]{\num{66.50}} & \cellcolor{gray!15}\makebox[0pt][c]{43.17} & \cellcolor{gray!15}\makebox[0pt][c]{73.09} & \cellcolor{gray!15}\makebox[0pt][c]{73.53} & \cellcolor{gray!15}\makebox[0pt][c]{68.19} & \cellcolor{gray!15}\makebox[0pt][c]{32.43} & \cellcolor{gray!15}\makebox[0pt][c]{56.12} & \cellcolor{gray!15}\makebox[0pt][c]{61.47} \\ 
        \hline
        \multirow{3}{*}{3.03} & GPTQ & 79.27 & 66.16 & 41.55 & 77.89 & 75.49 & 67.8 & 31.66 & 57.87 & 62.21 \\ 
        ~ & BSP & 79.68 & 65.11 & 42.32 & 70.73 & 75.36 & 66.38 & 30.79 & 56.65 & 60.88 \\ 
        ~ & \cellcolor{gray!15}\makebox[0pt][c]{EAC-MoE} & \cellcolor{gray!15}\makebox[0pt][c]{80.47} & \cellcolor{gray!15}\makebox[0pt][c]{\num{67.30}} & \cellcolor{gray!15}\makebox[0pt][c]{41.89} & \cellcolor{gray!15}\makebox[0pt][c]{77.99} & \cellcolor{gray!15}\makebox[0pt][c]{75.79} & \cellcolor{gray!15}\makebox[0pt][c]{69.22} & \cellcolor{gray!15}\makebox[0pt][c]{31.46} & \cellcolor{gray!15}\makebox[0pt][c]{58.97} & \cellcolor{gray!15}\makebox[0pt][c]{62.89} \\ 
        \hline
        \hline
    \end{tabular}}
\end{table*}

\subsection{Reproduction Details of Pruning}
\label{sec:repoduce-pruning}
EES \citep{lu-etal-2024-experts} and ODP \citep{huang2024mc} are two popular dynamic pruning methods for MoE models, both focusing on ignoring the least-contributing experts for each input token. ODP extends EES by incorporating a significance-aware token protection mechanism.

For EES, we use the same calibration dataset as in its paper to compute the ratio between the weight of the least-contributing expert and the weight of the most-contributing expert for each token. The median of all these ratios is selected as the pruning threshold. During inference, if the ratio of the least-contributing expert's weight to the most-contributing expert's weight for a given token is less than this threshold, the weight of the least-contributing expert is set to zero.

For ODP, we follow the same procedure as EES to determine the pruning threshold. On top of this, we incorporate the Significance-Aware Token Protection mechanism mentioned in the paper. This mechanism dynamically identifies critical tokens and prevents the pruning of the least-contributing experts for these critical tokens, even if they meet the original pruning condition.

\subsection{Completed Results of Pruning}
\label{sec:all-results-pruning}
In~\Cref{tab:quant-qwen1.5-moe-a2.7b-pruning}, we present the full results of ~\Cref{tab:speedup_comparison}. We compare our approach with several other methods, including EES \citep{lu-etal-2024-experts} and ODP \citep{huang2024mc}. 

\begin{table*}[!ht]
    \centering
    \caption{Comprehensive comparison of average accuracy of four models on 8 Zero-Shot tasks under pruning.}
    \label{tab:quant-qwen1.5-moe-a2.7b-pruning}
    \resizebox{\textwidth}{!}{
    \begin{tabular}{c|c|ccccccccccc}
    \hline
    \hline
        \parbox{2cm}{\centering Model} & Method & PIQA & ARC-E & ARC-C & BOOLQ & HS & WG & MATHQA & MMLU & Avg \\ 
        \hline
        \multirow{5}{*}{Mixtral-8x7B} & bf16 & 83.57 & 83.67 & \num{59.30} & 85.35 & 84.03 & 76.24 & 41.64 & 67.29 & 72.64 \\ 
        ~ & EES & \num{82.70} & \num{81.40} & 58.45 & 87.74 & 83.00 & 75.24 & 41.14 & 64.50 & \num{71.40} \\ 
        ~ & ODP & 82.87 & 83.21 & 59.01 & 84.98 & 83.69 & 75.97 & 40.93 & 65.17 & 71.98 \\
        ~ & \cellcolor{gray!15} PESF (\(\alpha=0.3\)) & \cellcolor{gray!15}\makebox[0pt][c]{83.19} & \cellcolor{gray!15}\makebox[0pt][c]{83.29} & \cellcolor{gray!15}\makebox[0pt][c]{59.90} & \cellcolor{gray!15}\makebox[0pt][c]{85.08} & \cellcolor{gray!15}\makebox[0pt][c]{\num{83.40}} & \cellcolor{gray!15}\makebox[0pt][c]{\num{76.40}} & \cellcolor{gray!15}\makebox[0pt][c]{40.90} & \cellcolor{gray!15}\makebox[0pt][c]{65.36} & \cellcolor{gray!15}\makebox[0pt][c]{72.19} \\ 
        ~ & \cellcolor{gray!30} PESF (\(\alpha=0.7\)) & \cellcolor{gray!30}\makebox[0pt][c]{68.23} & \cellcolor{gray!30}\makebox[0pt][c]{65.15} & \cellcolor{gray!30}\makebox[0pt][c]{45.99} & \cellcolor{gray!30}\makebox[0pt][c]{\num{78.10}} & \cellcolor{gray!30}\makebox[0pt][c]{69.39} & \cellcolor{gray!30}\makebox[0pt][c]{\num{64.80}} & \cellcolor{gray!30}\makebox[0pt][c]{29.25} & \cellcolor{gray!30}\makebox[0pt][c]{44.86} & \cellcolor{gray!30}\makebox[0pt][c]{58.22} \\ 
        \hline
        \hline
        \multirow{5}{*}{Phi3.5-moe} & bf16 & 77.69 & \num{65.70} & 53.58 & 88.35 & 79.87 & \num{76.80} & 38.32 & 76.62 & 69.62 \\ 
        ~ & EES & \num{76.50} & 63.85 & 51.37 & 86.94 & \num{78.60} & 75.14 & 36.25 & 75.02 & 67.96 \\ 
        ~ & ODP & 77.23 & 64.94 & 53.55 & 88.10 & 79.21 & 74.93 & 38.27 & 75.11 & 68.92 \\ 
        ~ & \cellcolor{gray!15} PESF (\(\alpha=0.3\)) & \cellcolor{gray!15}\makebox[0pt][c]{77.04} & \cellcolor{gray!15}\makebox[0pt][c]{65.53} & \cellcolor{gray!15}\makebox[0pt][c]{54.01} & \cellcolor{gray!15}\makebox[0pt][c]{\num{88.20}} & \cellcolor{gray!15}\makebox[0pt][c]{79.82} & \cellcolor{gray!15}\makebox[0pt][c]{75.30} & \cellcolor{gray!15}\makebox[0pt][c]{37.99} & \cellcolor{gray!15}\makebox[0pt][c]{\num{76.30}} & \cellcolor{gray!15}\makebox[0pt][c]{69.27} \\ 
        ~ & \cellcolor{gray!30} PESF (\(\alpha=0.7\)) & \cellcolor{gray!30}\makebox[0pt][c]{75.52} & \cellcolor{gray!30}\makebox[0pt][c]{64.94} & \cellcolor{gray!30}\makebox[0pt][c]{51.54} & \cellcolor{gray!30}\makebox[0pt][c]{86.91} & \cellcolor{gray!30}\makebox[0pt][c]{79.51} & \cellcolor{gray!30}\makebox[0pt][c]{73.16} & \cellcolor{gray!30}\makebox[0pt][c]{37.76} & \cellcolor{gray!30}\makebox[0pt][c]{74.24} & \cellcolor{gray!30}\makebox[0pt][c]{67.95} \\ 
        \hline
        \hline
        \multirow{5}{*}{Deepseek-moe-16b-base} & bf16 & 80.52 & 73.19 & 47.53 & 72.57 & 77.43 & 69.93 & 31.66 & 38.18 & 61.38 \\ 
        ~ & EES & 79.98 & 72.85 & 48.21 & 72.51 & 77.33 & 70.11 & 29.88 & 38.33 & 61.15 \\ 
        ~ & ODP & 80.01 & 72.94 & 47.41 & 73.39 & 77.25 & 70.14 & 30.23 & 38.14 & 61.19 \\ 
        ~ & \cellcolor{gray!15} PESF (\(\alpha=0.3\)) & \cellcolor{gray!15}\makebox[0pt][c]{80.52} & \cellcolor{gray!15}\makebox[0pt][c]{73.02} & \cellcolor{gray!15}\makebox[0pt][c]{46.93} & \cellcolor{gray!15}\makebox[0pt][c]{72.72} & \cellcolor{gray!15}\makebox[0pt][c]{77.13} & \cellcolor{gray!15}\makebox[0pt][c]{70.32} & \cellcolor{gray!15}\makebox[0pt][c]{31.52} & \cellcolor{gray!15}\makebox[0pt][c]{38.07} & \cellcolor{gray!15}\makebox[0pt][c]{61.28} \\ 
        ~ & \cellcolor{gray!30} PESF (\(\alpha=0.7\)) & \cellcolor{gray!30}\makebox[0pt][c]{78.45} & \cellcolor{gray!30}\makebox[0pt][c]{73.15} & \cellcolor{gray!30}\makebox[0pt][c]{\num{45.90}} & \cellcolor{gray!30}\makebox[0pt][c]{75.29} & \cellcolor{gray!30}\makebox[0pt][c]{75.68} & \cellcolor{gray!30}\makebox[0pt][c]{68.51} & \cellcolor{gray!30}\makebox[0pt][c]{31.49} & \cellcolor{gray!30}\makebox[0pt][c]{34.83} & \cellcolor{gray!30}\makebox[0pt][c]{60.41} \\ 
        \hline
        \hline
        \multirow{5}{*}{Qwen1.5-MoE-A2.7B} & bf16 & 80.79 & 69.44 & 44.37 & 79.57 & 77.17 & 69.77 & 35.57 & 61.08 & 64.72 \\ 
        ~ & EES & 80.52 & 69.11 & 44.01 & 79.45 & 76.87 & 69.32 & 35.11 & 60.97 & 64.42 \\ 
        ~ & ODP & 80.61 & 69.24 & 44.12 & 79.51 & 77.01 & 69.32 & 35.05 & 60.95 & 64.48 \\ 
        ~ & \cellcolor{gray!15} PESF (\(\alpha=0.3\)) & \cellcolor{gray!15}\makebox[0pt][c]{80.74} & \cellcolor{gray!15}\makebox[0pt][c]{69.28} & \cellcolor{gray!15}\makebox[0pt][c]{44.28} & \cellcolor{gray!15}\makebox[0pt][c]{79.36} & \cellcolor{gray!15}\makebox[0pt][c]{77.24} & \cellcolor{gray!15}\makebox[0pt][c]{69.77} & \cellcolor{gray!15}\makebox[0pt][c]{35.68} & \cellcolor{gray!15}\makebox[0pt][c]{\num{60.80}} & \cellcolor{gray!15}\makebox[0pt][c]{64.64} \\ 
        ~ & \cellcolor{gray!30} PESF (\(\alpha=0.7\)) & \cellcolor{gray!30}\makebox[0pt][c]{80.25} & \cellcolor{gray!30}\makebox[0pt][c]{69.11} & \cellcolor{gray!30}\makebox[0pt][c]{43.94} & \cellcolor{gray!30}\makebox[0pt][c]{77.06} & \cellcolor{gray!30}\makebox[0pt][c]{76.44} & \cellcolor{gray!30}\makebox[0pt][c]{70.09} & \cellcolor{gray!30}\makebox[0pt][c]{35.38} & \cellcolor{gray!30}\makebox[0pt][c]{58.72} & \cellcolor{gray!30}\makebox[0pt][c]{63.87} \\ 
        \hline
        \hline
    \end{tabular}}
\end{table*}

\subsection{Completed Results of QESC+PESF}
\label{sec:all-results-EAC-MOE}
In this subsection, leveraging the three quantization bit-width settings from QESC and the two pruning threshold strategies from PESF, we provide a detailed evaluation of different combinations on four models. The results include the average zero-shot accuracy and inference speedup ratio, along with specific accuracy on each dataset.

\noindent\textbf{Detailed results.}
Here, we provide detailed results of our method under three quantization bit-widths and two pruning thresholds, as shown in ~\Cref{tab:EAC-MOE-OVERALL}. Furthermore, as shown in ~\Cref{tab:EAC-MOE-mixtral-8x7B,tab:EAC-MOE-Phi3.5-moe,tab:EAC-MOE-Deepseek-moe,tab:EAC-MOE-Qwen1.5-moe}, we report the specific results for various configurations of the four models on each dataset. "EAC-MoE" represents the use of conservative PESF pruning strategy (\(\alpha=0.3\)) based on QESC, while "EAC-MoE$^*$" represents the use of a more aggressive PESF pruning strategy (\(\alpha=0.7\)) based on QESC.

\begin{table*}[ht]
    \caption{Comparison of the average accuracy on 8 zero-shot tasks and speedup of inference across four MoE models under quantization (QESC) and pruning (PESF). "EAC-MoE" represents the use of conservative PESF pruning strategy (\(\alpha=0.3\)) based on QESC, "EAC-MoE$^*$" represents the use of PESF (\(\alpha=0.7\)) based on QESC.}
    \label{tab:EAC-MOE-OVERALL}
    \centering
    \resizebox{\textwidth}{!}{
    \begin{tabular}{c|c||cc|cc|cc|cc}
        \hline
        \hline
        \multirow{2}{*}{\parbox{2cm}{\centering Bits}} & \multirow{2}{*}{\parbox{3cm}{\centering Method}} & \multicolumn{2}{c|}{Mixtral-8x7B} & \multicolumn{2}{c|}{Phi3.5-moe} & \multicolumn{2}{c|}{Deepseek-moe-16b-base} & \multicolumn{2}{c}{Qwen1.5-MoE-A2.7B} \\ 
        ~ & ~ & $\text{0-shot}^{8} \uparrow$ & Speedup $\uparrow$ & $\text{0-shot}^{8} \uparrow$ & Speedup $\uparrow$ & $\text{0-shot}^{8} \uparrow$ & Speedup $\uparrow$ & $\text{0-shot}^{8} \uparrow$ & Speedup $\uparrow$ \\
        \hline
        \multirow{1}{*}{16.00} & Baseline & 72.64 & 1.00 & 69.62 & 1.00 & 61.38 & 1.00 & 64.72 & 1.00 \\ 
        \hline
        \multirow{2}{*}{2.06} 
        & EAC-MoE & 65.90 & 1.82 & 65.26 & 1.84 & 56.75 & 1.59 & 61.18 & 1.64 \\
        ~ & EAC-MoE$^*$ & 49.09 & 1.93 & 62.16 & 2.03 & 54.81 & 2.01 & 60.02 & 2.08 \\
        \hline
        \multirow{2}{*}{2.54} 
        & EAC-MoE & 68.60 & 1.74 & 65.98 & 1.80 & 58.17 & 1.56 & 61.41 & 1.60 \\ 
        ~ & EAC-MoE$^*$ & 53.01 & 1.81 & 63.80 & 2.02 & 56.83 & 1.98 & 60.57 & 2.03 \\ 
        \hline
        \multirow{2}{*}{3.03} 
        & EAC-MoE & 71.68 & 1.68 & 68.31 & 1.75 & 61.09 & 1.55 & 62.73 & 1.58 \\
        ~ & EAC-MoE$^*$ & 57.55 & 1.76 & 66.66 & 1.98 & 58.82 & 1.96 & 61.77 & 2.01 \\
        \hline
        \hline
    \end{tabular}}
\end{table*}

\begin{table*}[!ht]
    \centering
    \caption{Detailed results of average accuracy of Mixtral-8x7B on 8 Zero-Shot tasks under quantization and pruning.}
    \label{tab:EAC-MOE-mixtral-8x7B}
    \resizebox{\textwidth}{!}{
    \begin{tabular}{c|l|ccccccccccc}
    \hline
    \hline
        \parbox{2cm}{\centering Bits} & Method & PIQA & ARC-E & ARC-C & BOOLQ & HS & WG & MATHQA & MMLU & Avg \\ 
        \hline
        \multirow{1}{*}{16.00} & Full Precision & 83.57 & 83.67 & 59.3 & 85.35 & 84.03 & 76.24 & 41.64 & 67.29 & 72.64 \\ 
        \hline
        \multirow{2}{*}{2.06} & EAC-MoE & 79.76 & 77.78 & 50.85 & 77.28 & 75.80 & 73.24 & 35.88 & 56.60 & 65.90 \\ 
        ~ & EAC-MoE$^*$ & 61.04 & 54.42 & 38.23 & 69.54 & 48.15 & 61.09 & 24.69 & 35.58 & 49.09 \\ 
        \hline
        \multirow{2}{*}{2.54} & EAC-MoE & 81.28 & 80.13 & 54.44 & 83.85 & 78.66 & 74.27 & 37.39 & 58.78 & 68.60 \\ 
        ~ & EAC-MoE$^*$ & 65.29 & 58.63 & 39.25 & 75.05 & 58.27 & 61.56 & 27.10 & 38.92 & 53.01 \\ 
        \hline
        \multirow{2}{*}{3.03} & EAC-MoE & 83.03 & 82.91 & 58.28 & 85.35 & 82.17 & 76.87 & 41.47 & 63.35 & 71.68 \\ 
        ~ & EAC-MoE$^*$ & 69.98 & 64.01 & 42.04 & 81.92 & 67.24 & 63.22 & 29.98 & 42.01 & 57.55 \\ 
        \hline
        \hline
    \end{tabular}}
\end{table*}

\begin{table*}[!ht]
    \centering
    \caption{Detailed results of average accuracy of Phi3.5-moe on 8 Zero-Shot tasks under quantization and pruning.}
    \label{tab:EAC-MOE-Phi3.5-moe}
    \resizebox{\textwidth}{!}{
    \begin{tabular}{c|l|ccccccccccc}
    \hline
    \hline
        \parbox{2cm}{\centering Bits} & Method & PIQA & ARC-E & ARC-C & BOOLQ & HS & WG & MATHQA & MMLU & Avg \\ 
        \hline
        \multirow{1}{*}{16.00} & Full Precision & 77.69 & 65.70 & 53.58 & 88.35 & 79.87 & 76.80 & 38.32 & 76.62 & 69.62 \\ 
        \hline
        \multirow{2}{*}{2.06} & EAC-MoE & 74.16 & 61.28 & 48.55 & 86.64 & 75.41 & 73.56 & 33.10 & 69.35 & 65.26 \\ 
        ~ & EAC-MoE$^*$ & 72.80 & 59.01 & 49.40 & 83.27 & 72.74 & 69.30 & 32.09 & 58.68 & 62.16 \\ 
        \hline
        \multirow{2}{*}{2.54} & EAC-MoE & 75.73 & 60.61 & 48.72 & 86.64 & 76.84 & 72.69 & 35.44 & 71.18 & 65.98 \\ 
        ~ & EAC-MoE$^*$ & 73.01 & 58.59 & 48.38 & 84.37 & 74.10 & 68.67 & 34.44 & 68.82 & 63.80 \\ 
        \hline
        \multirow{2}{*}{3.03} & EAC-MoE & 76.93 & 62.38 & 52.39 & 88.13 & 78.38 & 76.01 & 36.65 & 75.57 & 68.31 \\ 
        ~ & EAC-MoE$^*$ & 74.65 & 61.28 & 50.68 & 86.24 & 77.77 & 72.77 & 36.75 & 73.17 & 66.66 \\ 
        \hline
        \hline
    \end{tabular}}
\end{table*}

\begin{table*}[!ht]
    \centering
    \caption{Detailed results of average accuracy of Deepseek-moe-16b-base on 8 Zero-Shot tasks under quantization and pruning.}
    \label{tab:EAC-MOE-Deepseek-moe}
    \resizebox{\textwidth}{!}{
    \begin{tabular}{c|l|ccccccccccc}
    \hline
    \hline
        \parbox{2cm}{\centering Bits} & Method & PIQA & ARC-E & ARC-C & BOOLQ & HS & WG & MATHQA & MMLU & Avg \\ 
        \hline
        \multirow{1}{*}{16.00} & Full Precision & 80.52 & 73.19 & 47.53 & 72.57 & 77.43 & 69.93 & 31.66 & 38.18 & 61.37 \\ 
        \hline
        \multirow{2}{*}{2.06} & EAC-MoE & 77.04 & 69.11 & 41.30 & 72.55 & 69.85 & 67.48 & 28.61 & 28.07 & 56.75 \\ 
        ~ & EAC-MoE$^*$ & 75.19 & 67.93 & 39.93 & 70.45 & 66.55 & 63.85 & 27.77 & 26.81 & 54.81 \\ 
        \hline
        \multirow{2}{*}{2.54} & EAC-MoE & 78.29 & 69.15 & 42.32 & 74.50 & 66.55 & 63.85 & 27.77 & 26.81 & 54.81 \\ 
        ~ & EAC-MoE$^*$ & 76.17 & 71.74 & 41.13 & 71.74 & 68.31 & 66.85 & 28.11 & 30.57 & 56.83 \\ 
        \hline
        \multirow{2}{*}{3.03} & EAC-MoE & 79.54 & 73.15 & 46.16 & 75.02 & 75.55 & 70.24 & 31.59 & 37.45 & 61.09 \\ 
        ~ & EAC-MoE$^*$ & 77.20 & 71.55 & 45.05 & 72.51 & 72.86 & 66.46 & 31.56 & 33.33 & 58.82 \\ 
        \hline
        \hline
    \end{tabular}}
\end{table*}

\begin{table*}[!ht]
    \centering
    \caption{Detailed results of average accuracy of Qwen1.5-MoE-A2.7B on 8 Zero-Shot tasks under quantization and pruning.}
    \label{tab:EAC-MOE-Qwen1.5-moe}
    \resizebox{\textwidth}{!}{
    \begin{tabular}{c|l|ccccccccccc}
    \hline
    \hline
        \parbox{2cm}{\centering Bits} & Method & PIQA & ARC-E & ARC-C & BOOLQ & HS & WG & MATHQA & MMLU & Avg \\ 
        \hline
        \multirow{1}{*}{16.00} & Full Precision & 80.79 & 69.44 & 44.37 & 79.57 & 77.17 & 69.77 & 35.57 & 61.08 & 64.72 \\ 
        \hline
        \multirow{2}{*}{2.06} & EAC-MoE & 79.16 & 65.07 & 42.83 & 75.01 & 72.44 & 68.19 & 32.63 & 54.12 & 61.18 \\ 
        ~ & EAC-MoE$^*$ & 77.91 & 64.52 & 42.15 & 74.86 & 70.94 & 67.09 & 32.26 & 50.46 & 60.02 \\ 
        \hline
        \multirow{2}{*}{2.54} & EAC-MoE & 78.78 & 66.75 & 43.17 & 72.57 & 73.53 & 68.19 & 32.29 & 56.01 & 61.41 \\ 
        ~ & EAC-MoE$^*$ & 77.53 & 65.24 & 40.44 & 74.89 & 72.49 & 68.27 & 32.16 & 53.52 & 60.57 \\ 
        \hline
        \multirow{2}{*}{3.03} & EAC-MoE & 80.41 & 66.92 & 42.15 & 76.88 & 75.81 & 69.22 & 31.72 & 58.72 & 62.73 \\ 
        ~ & EAC-MoE$^*$ & 79.16 & 66.71 & 41.30 & 75.35 & 74.97 & 68.75 & 31.99 & 55.96 & 61.77 \\ 
        \hline
        \hline
    \end{tabular}}
\end{table*}

\subsection{Task-Preference and Sparsity in Experts-Selection}
\label{sec:sparsity_in_experts}

\begin{figure*}[ht]
    \centering
    \includegraphics[width=\textwidth]{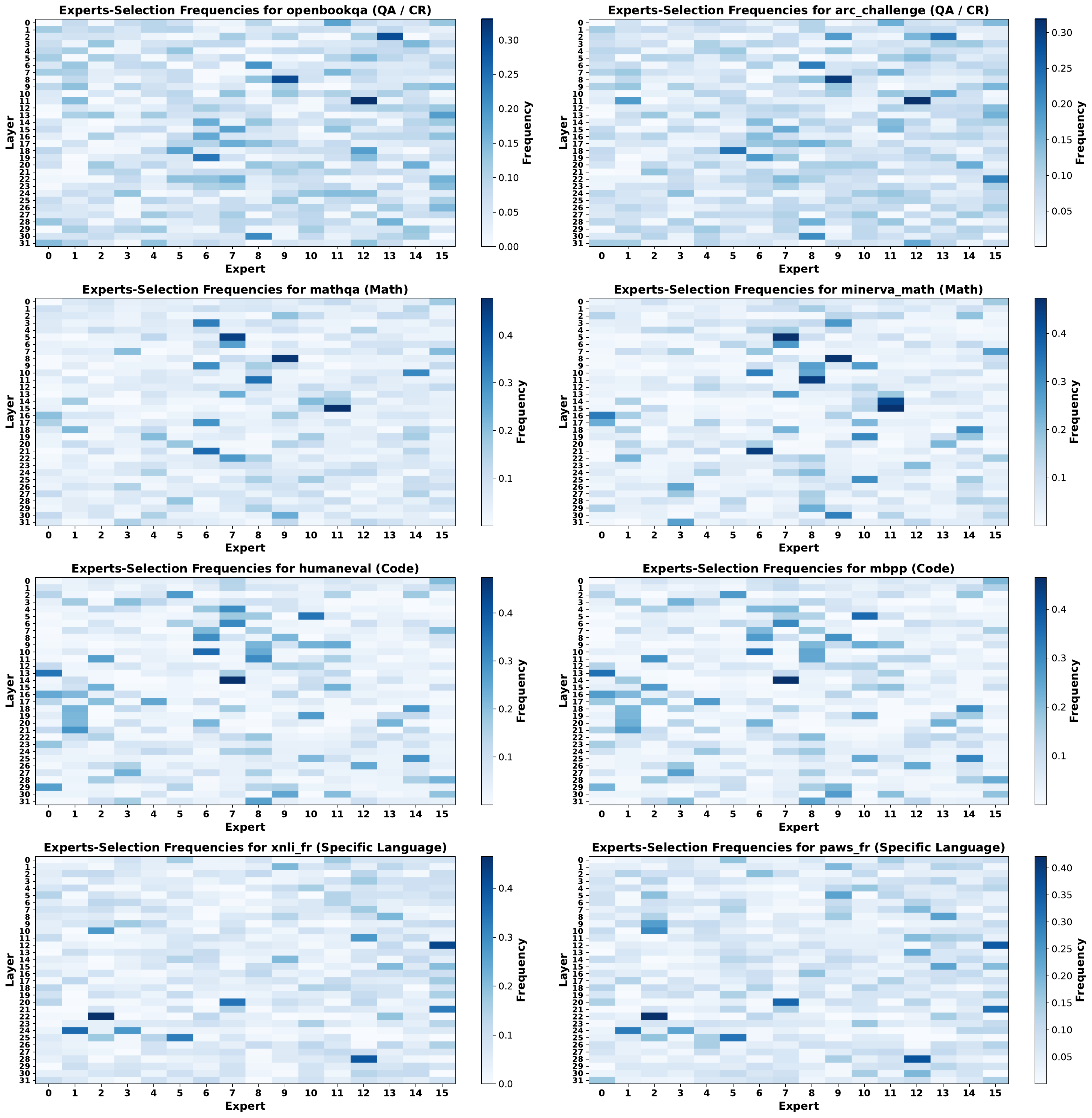}
    \caption{The frequency of expert selection across 8 datasets spanning 4 task types for Phi3.5-moe.}
    \label{fig:appendix_phi_freq_all}
\end{figure*}

\begin{figure*}[ht]
    \centering
    \includegraphics[width=\textwidth]{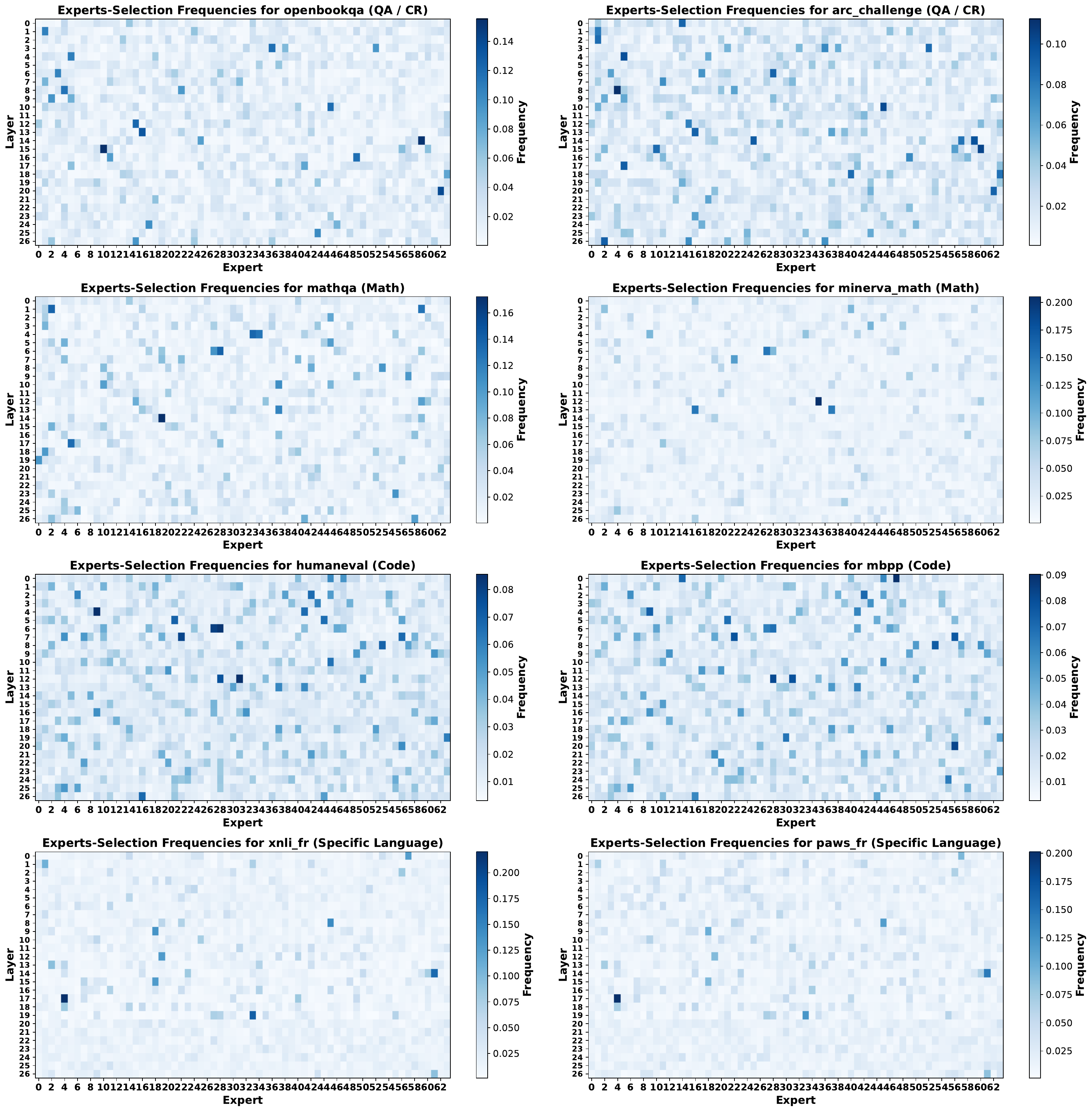}
    \caption{The frequency of expert selection across 8 datasets spanning 4 task types for Deepseek-moe-16b-base}
    \label{fig:appendix_dps_freq_all}
    \vskip -0.2in
\end{figure*}

In ~\Cref{sec:generalization}, we calculate the expert selection frequency and pairwise cosine similarity of MoE models across different datasets for four types of reasoning tasks, providing a macroscopic view of the task preferences in expert selection. In this section, we delve deeper into the microscopic details, specifically discussing the expert selection preferences of the Phi3.5-moe and Deepseek-moe-16b-base models across 8 datasets spanning 4 task types. This analysis also aligns with previous work \citep{li2024examining}, further demonstrating their sparsity.
As shown in ~\Cref{fig:appendix_phi_freq_all}, the four rows, from top to bottom, correspond to four different task types ((QA/CR), Math, Code, Specific Language). For each row, the left and right plots respectively show the expert selection frequencies of Phi3.5-moe on two different datasets of the same task type. From this, a clear pattern can be observed, indicating that the MoE model exhibits remarkably similar expert selection preferences across datasets within the same task category. For example, in the first row, as illustrated in the left and right subfigures, certain experts such as Expert13 in Layer2, Expert9 in Layer8, and Expert12 in Layer11 are frequently selected for the openbookqa and arc-challenge datasets, with average selection frequencies exceeding 30\% (note that each layer has 16 experts, and a completely balanced selection would result in a frequency of 6.25\% per expert). Conversely, some experts, such as Expert1 in Layer5, Expert7 in Layer14, and Expert8 in Layer27, are rarely selected, with frequencies below 1\%.

Similarly, in the second row, for the Math task across two datasets, experts such as Expert7 in Layer5, Expert9 in Layer8, and Expert11 in Layer15 are selected with an average frequency exceeding 40\%, while others, such as Expert0 in Layer0 and Expert8 in Layer2, are seldom chosen. These results provide detailed evidence that Phi3.5-moe demonstrates a high degree of similarity in expert selection frequencies within task categories while also exhibiting significant sparsity. From another perspective, Phi3.5-moe demonstrates entirely distinct expert selection preferences across different reasoning tasks. For instance, Expert13 in Layer2 is frequently selected for (QA/CR) tasks but is neither prominent nor frequently chosen in the other three tasks. Similarly, Expert7 in Layer14 is heavily utilized in Code tasks but is rarely selected in the other three task categories.

A similar pattern is observed in Deepseek-moe-16b-base, which has 64 experts per layer, as shown in ~\Cref{fig:appendix_dps_freq_all}. While displaying clear intra-category similarities and inter-category differences in expert selection, the larger number of experts results in an even greater degree of sparsity in expert selection for Deepseek-moe-16b-base.

\subsection{Pruning on Mixtral-8x7B}
\label{sec:pruning-mixtral-8x7B}
\begin{figure}[ht]
    \centering
    \includegraphics[width=\linewidth]{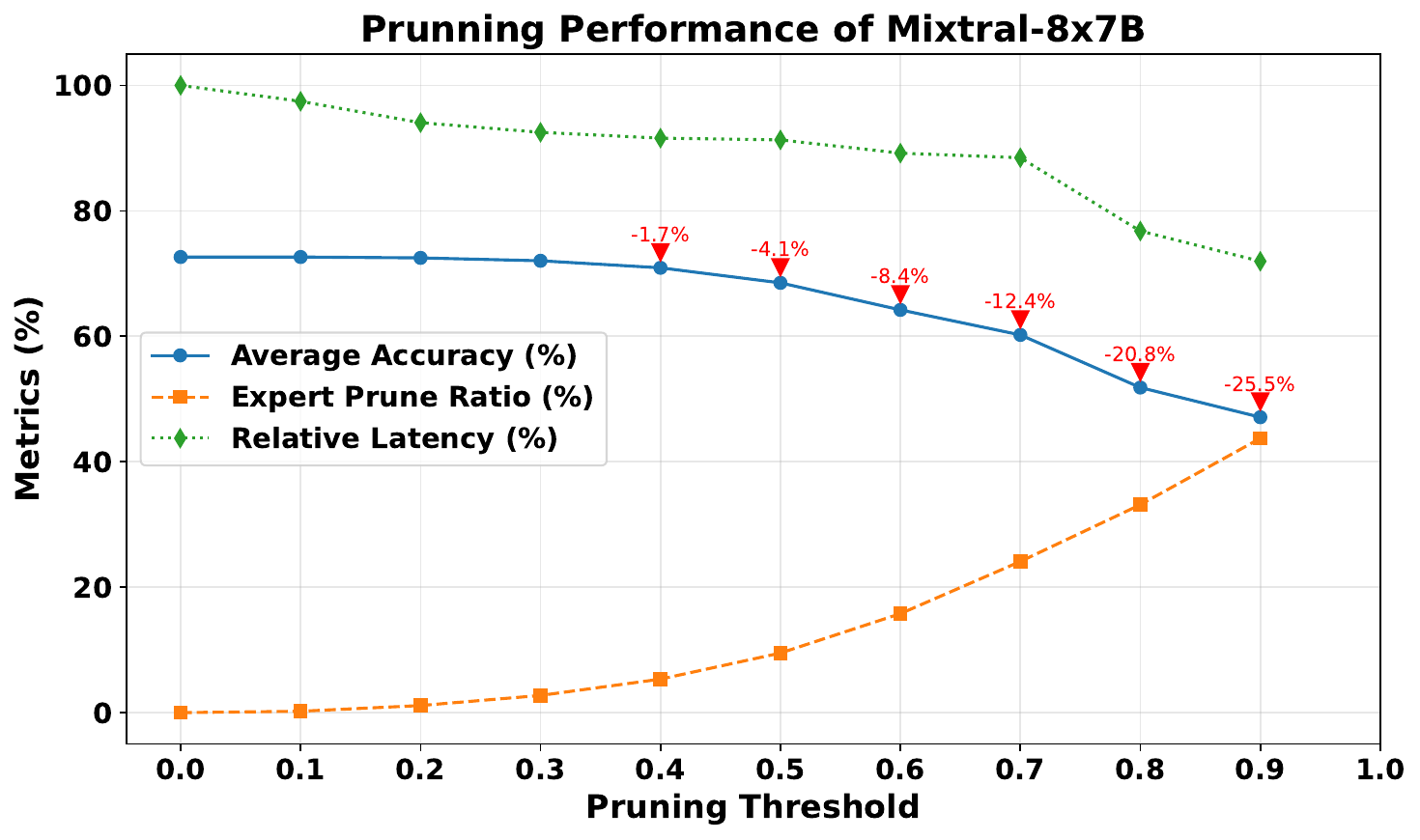}
    \caption{The variations in the model’s average accuracy, expert pruning rate, and inference acceleration effect with respect to changes in the pruning threshold for Mixtral-8x7B.}
    \label{fig:pruning-mixtal-8x7B}
\end{figure}


\begin{figure}[htbp] 
    \centering
    \begin{subfigure}[b]{\linewidth}
        \centering
        \includegraphics[width=0.8\linewidth]{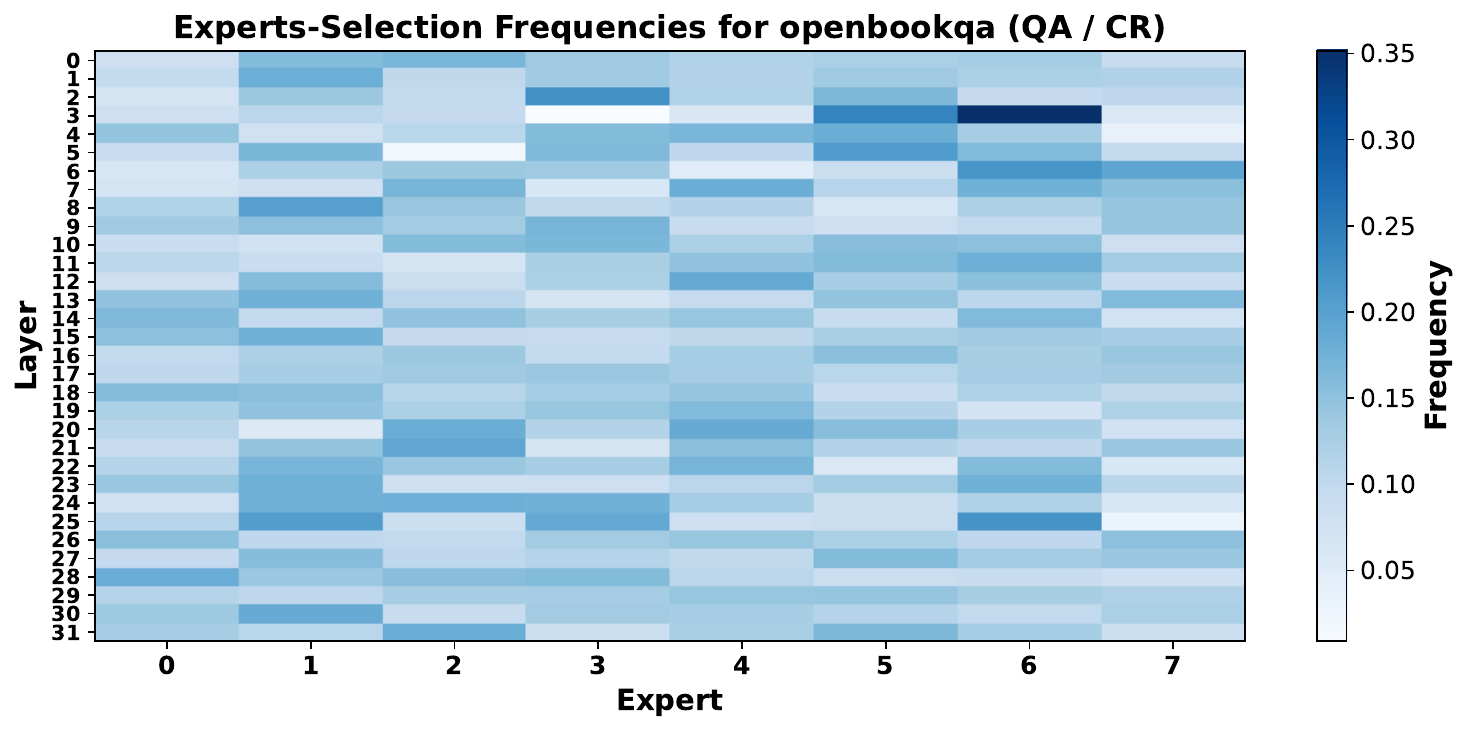} 
        \label{fig:mixtral-1}
    \end{subfigure}


    \begin{subfigure}[b]{\linewidth}
        \centering
        \includegraphics[width=0.8\linewidth]{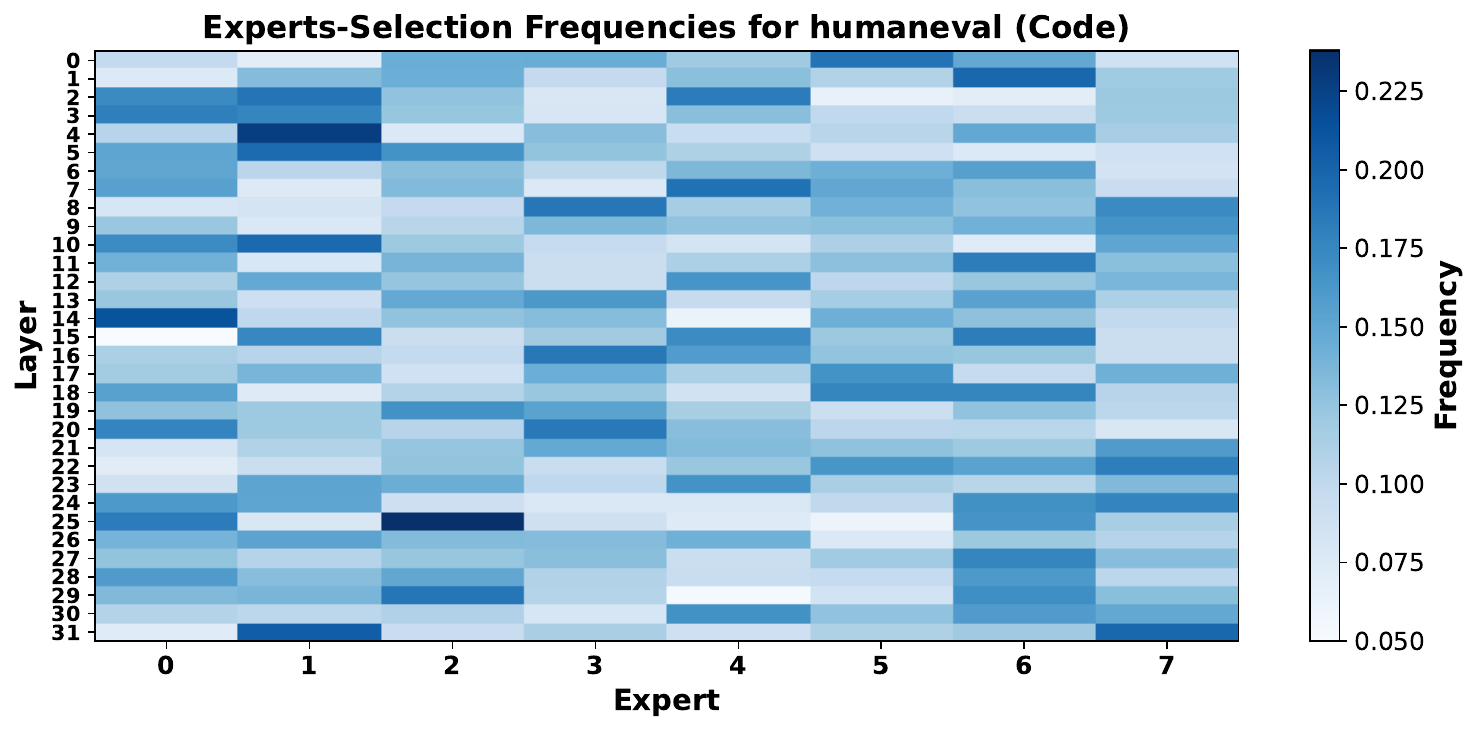} 
        \label{fig:mixtral-2}
    \end{subfigure}
    \caption{The frequency of expert selection on openbookqa and humaneval for Mixtral-8x7B.}
    \label{fig:mixtral-freq}
\end{figure}

In ~\Cref{sec:exp-pruning}, when employing a more aggressive pruning strategy, unlike the other three models which maintain relatively stable average accuracy under significant inference speedup, Mixtral-8x7B exhibits notable performance degradation. This section delves into this phenomenon and analyzes its underlying causes.

Similar to ~\Cref{fig:pruning-mixtal-8x7B}, we plot the changes in Mixtral-8x7B's average accuracy, expert pruning rate, and inference speedup as the pruning threshold varied. As shown in the figure, unlike Phi3.5-moe and Deepseek-moe-16b-base, where a significant drop in accuracy only occurs when the pruning threshold exceeded 0.7 and the expert pruning rate approached 40\%, Mixtral-8x7B begins to show a noticeable decline in accuracy once the pruning threshold surpassed 0.3.

We further analyze the expert selection frequency of Mixtral-8x7B across two different datasets. As illustrated in ~\Cref{fig:mixtral-freq}, compared to Phi3.5-moe and Deepseek-moe-16b-base, Mixtral-8x7B exhibits weaker sparsity in expert selection. Apart from a few experts, such as Expert6 in Layer3 (top) and Expert2 in Layer25 (bottom), whose average selection frequencies exceed the mean (0.125), the selection frequencies of the remaining experts are relatively balanced. This phenomenon has also been noted in \citep{jiang2024mixtral, li2024examining}. Consequently, Mixtral-8x7B is more sensitive to dynamic expert pruning compared to the other three models, making it less suitable for the aggressive pruning settings (\(\alpha=0.7\)) proposed in our PESF method. Nevertheless, our approach achieved commendable inference speedup and accuracy retention on Mixtral-8x7B under conservative pruning settings (\(\alpha=0.3\)) . In the future, we aim to explore methods to achieve higher pruning rates and speedup while maintaining accuracy on Mixtral-8x7B.

\subsection{Datasets used in Expert-Selection Analysis}
\label{sec:datasets_analysis}
In ~\Cref{sec:generalization}, we record Expert-Selection frequency on 15 datasets of three common categories of NLP tasks——Math, Code-Generation, Question-Answering or Commonsense-Reasoning (QA/CR) and 4 datasets in French language.
GSM8K \citep{cobbe2021gsm8k}, MathQA \citep{amini-etal-2019-mathqa}, Minerva\_Math \citep{NEURIPS2022_18abbeef} and Hendrycks\_Math \citep{hendrycksmath2021} are in Math task; Winogrande \citep{ai2:winogrande}, PIQA \citep{Bisk2020}, ARC-Challenge \citep{allenai:arc}, BoolQ \citep{clark2019boolq}, MathQA \citep{amini-etal-2019-mathqa}, HellaSwag \citep{zellers2019hellaswag}, and Social\_iqa \citep{sap-etal-2019-social} are in QA/CR task; Humaneval \citep{chen2021evaluating}, Mbpp \citep{austin2021program}, Apps \citep{hendrycksapps2021} and Conala \citep{yin2018learning} are in code task; Lambada\_fr \citep{paperno-EtAl:2016:P16-1}, Xnli\_fr \citep{conneau2018xnli}, Paws\_fr \citep{paws2019naacl} and Arc\_fr \citep{allenai:arc} are datasets in French language.

\end{document}